\documentclass[letterpaper, 10 pt, conference]{ieeeconf}  

\IEEEoverridecommandlockouts                              

\usepackage[hidelinks]{hyperref}
\usepackage{graphicx}
\usepackage{amsmath}
\usepackage{amsfonts}
\usepackage{amssymb}
\usepackage{todonotes}
\usepackage{multirow}
\usepackage{bbding} 
\usepackage{adjustbox}
\usepackage{colortbl}
\usepackage{xcolor}  
\usepackage{pgfplots}
\usepackage{subcaption}

\let\labelindent\relax
\usepackage{enumitem}

\title{\LARGE \bf
Object-centric Cross-modal Feature Distillation for Event-based Object Detection
}

\author{Lei Li$^{1,2}$, Alexander Linger$^{1}$, Mario Millhäusler$^{1}$, Vagia Tsiminaki$^{1}$, Yuanyou Li$^{1}$ and Dengxin Dai$^{1}$
\thanks{*Work done during internship at Huawei Zurich Research Center (ZRC). }
\thanks{$^{1}$ Huawei ZRC, Switzerland. {\tt\scriptsize {firstname.lastname}@huawei.com}}%
\thanks{$^{2}$ ETH Zurich, Switzerland. {\tt\scriptsize leilil@ethz.ch}}%
}

\begin{document}

\maketitle
\thispagestyle{empty}
\pagestyle{empty}

\begin{abstract}

Event cameras are gaining popularity due to their unique properties, such as their low latency and high dynamic range. One task where these benefits can be crucial is real-time object detection. However, RGB detectors still outperform event-based detectors due to the sparsity of the event data and missing visual details. In this paper, we develop a novel knowledge distillation approach to shrink the performance gap between these two modalities. To this end, we propose a cross-modality object detection distillation method that by design can focus on regions where the knowledge distillation works best. We achieve this by using an object-centric slot attention mechanism that can iteratively decouple features maps into object-centric features and corresponding pixel-features used for distillation. We evaluate our novel distillation approach on a synthetic and a real event dataset with aligned grayscale images as a teacher modality. We show that object-centric distillation allows to significantly improve the performance of the event-based student object detector, nearly halving the performance gap with respect to the teacher.

\end{abstract}

\section{Introduction}
Event cameras asynchronously measure brightness changes at independent pixel~\cite{gallego2020event}, which stands in contrast to standard cameras that measure the brightness within a fixed time interval. 
These two measurement principles result in two sensors with different characteristics and advantages. 
Event cameras due to their asynchronous nature have sub-millisecond latency, no motion blur, and a high dynamic range (up to 140dB)~\cite{gallego2020event}. 
In contrast, regular cameras can measure the absolute intensity information, resulting in richer more detailed images. 
Given how close the two modalities are in their representation, both are based on pixels, the question naturally arises how can we leverage one modality to improve the other?

\begin{figure}[htbp]
\centering
\scalebox{0.9}{
\begin{tabular}{cc}
\includegraphics[width=0.2\textwidth]{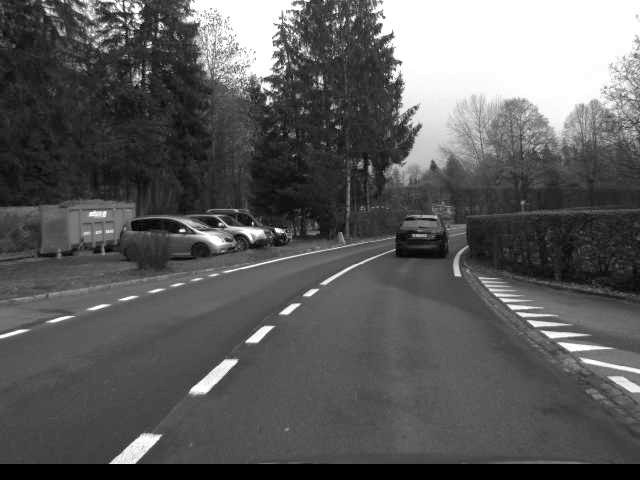}&
\includegraphics[width=0.2\textwidth]{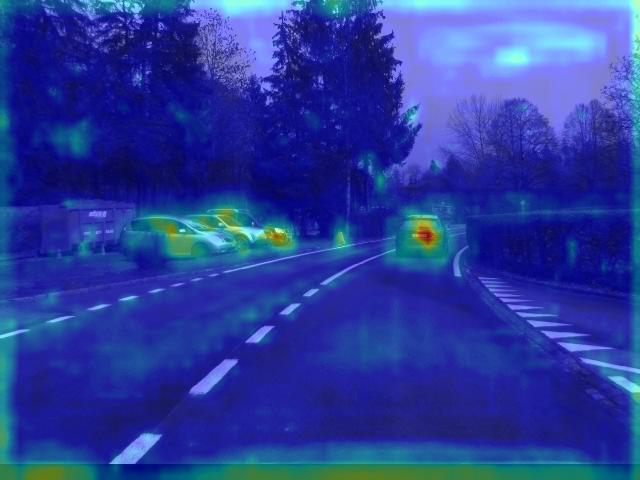}\\
Grayscale Input    &Grayscale Heatmap\\
\includegraphics[width=0.2\textwidth]{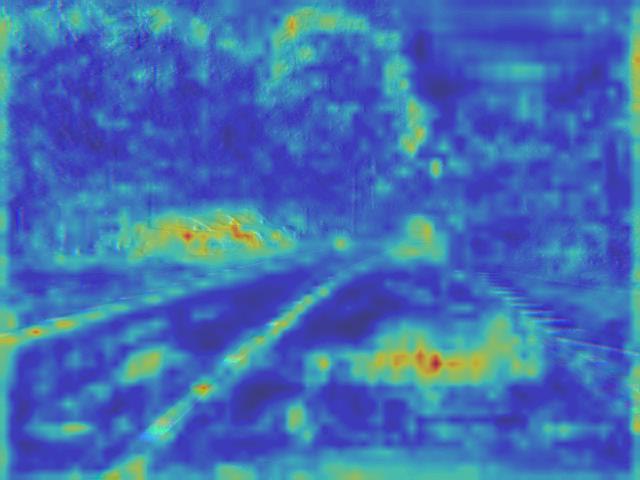}&
\includegraphics[width=0.2\textwidth]{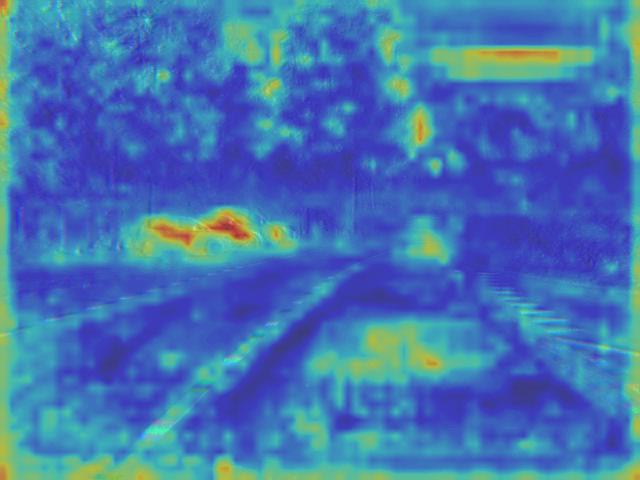} \\
Events Heatmap   &Events Heatmap (+Ours)\\

\vspace{-0.45cm}
\end{tabular}
}
\caption{Visualization of the spatial heat maps of the largest scale FPN feature in the YOLOX detector \cite{ge2021yolox} using grayscale image and event frame as inputs. 
%
%
With our distillation model, foreground features are enhanced and background noise (lane markings) is weakened.}
\vspace{-1.2em}
\label{fig:heatmap}
\end{figure}

In this paper, we study the task of object detection with event cameras~\cite{perot2020learning}. 
This is a task where the low latency of event cameras can help with fast moving objects, which helps to drastically improve reaction times in safety critical tasks such as autonomous driving. 
However, object detection with event cameras also has disadvantages, first, datasets for event camera based detection~\cite{perot2020learning} are not as mature as those for frame based cameras~\cite{geiger2013vision, caesar2020nuscenes, sun2020scalability, lin2014microsoft}. 
Second, the missing global illumination information results in inferior appearance features. 
Given these two disadvantages, frame based detectors generally perform better than event based detectors. 
To reduce this performance gap and maintain the property of event cameras, we investigate how cross-modal knowledge distillation during training can help to improve event-based detectors. 
We will show that this results in increased performance \emph{without} the need for frame-based information at inference time.

Knowledge distillation~\cite{hinton2015distilling} is one of the most common tools to transfer knowledge between models. 
However, this is often done from a heavy to a lighter model~\cite{hinton2015distilling, guo2021distilling, dai2021general, yang2022focal, yang2022masked, kang2021instance, nguyen2022improving}. 
We are interested in transferring knowledge between input modalities, which comes with its own set of challenges~\cite{chong2022monodistill,hong2022cross,chen2022bevdistill}. 
For cross modality knowledge distillation from event to frame cameras the main issue is that the event data is much sparser, thus it does not help to distill the information of the whole image but we propose to rather focus on the foreground objects that should be detected. 
The reason to focus on foreground objects can be directly understood when investigating the differences between the features generated from a detector using either events or image frames, shown as a heatmap in Fig.~\ref{fig:heatmap}.
It is clearly visible that event frames and grayscale images exhibit different pixel-level reactions. 
In event frames, objects appear as boundaries which is harder to distinguish from the background, while in RGB/grayscale images, they are displayed as shapes and textures. 
Thus, to distillate features between event frames and images, it is important to avoid to imitating the whole teacher's features directly at the pixel level. 
An additional problem is that due to the ego-motion, background areas can trigger a large number of events, additionally complicating the distillation of background regions. 

Thus, to get the best features to distill, we propose a novel distillation approach that uses slot attention~\cite{locatello2020object} to iteratively decouple feature maps into object-centric features and corresponding pixel features that allows to best distill the knowledge from the frame based teacher to the event based student. 
Our algorithm aligns the Feature Pyramid Network~\cite{lin2017feature} (FPN) output features between the student and the teacher and is thus applicable to a large number of modern detection algorithms. 
We show that our algorithm helps both one-stage (YOLOX~\cite{ge2021yolox}) and two-stage detectors (Faster R-CNN~\cite{ren2015faster}).

In summary, our contributions are threefold. 
First, we propose a novel slot attention based feature distillation algorithm that can iteratively decouple feature maps into object-centric features and corresponding pixel features for distillation. 
Second, we show that the attention matrices from the slot attention mechanism can be used as learning based mask for attention aided feature alignment and that considering the relation information between object-centric features can be exploited during distillation. 
Third, we show that the proposed distillation algorithm allows for improving the performance of event based detection algorithms, halving the performance gap to grayscale detectors.
\section{Related Work}
\subsection{Object Detection}
The addition of deep neural networks has significantly improved object detection over the last decade. 
Modern detectors are broadly classified into one-stage and two-stage detectors. One-stage detectors~\cite{law2018cornernet, zhou2019objects, redmon2016you, redmon2017yolo9000, redmon2018yolov3, ge2021yolox} generate predictions directly on the feature map and generally come with the advantage of faster run times. 
In contrast, two-stage detectors~\cite{ren2015faster, cai2018cascade} typically employ a region proposal network to generate coarse object bounding boxes, which are then refined in a second stage. 
Recently, major advances in object detection have focused on anchor-free detectors~\cite{law2018cornernet, zhou2019objects, ge2021yolox}, label assignment strategies~\cite{zhang2020bridging} and end-to-end detectors~\cite{carion2020end}. 

In the sub-field of real-time detection, the engineering-heavy YOLO family \cite{redmon2018yolov3, bochkovskiy2020yolov4, ge2021yolox, li2022yolov6} integrates recent advancements in detection technologies, always seeking the best speed and accuracy trade-offs for real-time applications. 
Due to the low latency of event cameras, these types of detectors are of special interest, as they allow to generate very high rate predictions if combined with asynchronous event cameras. 
In this paper, we utilize YOLOX~\cite{ge2021yolox}, which uses an anchor-free detection head and a sophisticated label assignment strategy.

By using a standard detector based on a Convolutional Neural Network (CNN), we follow the most common approach for event based object detection~\cite{perot2020learning, gehrig2022recurrent, chen2018pseudo, iacono2018towards}. 
Identical to ours, the idea is to first represent the event stream as a dense ``frame", which then can be processed by a CNN. These methods are often coupled with recurrent network architectures~\cite{perot2020learning, li2022asynchronous,gehrig2022recurrent} to deal with the sparse event data. 
An alternative approach to dense CNNs, is to process the events asynchronously using Graph Neural Networks (GNN) to achieve incremental detection~\cite{Li_2021_ICCV,schaefer2022aegnn, gehrig2022pushing}, which is more aligned with the characteristics of event cameras. 
However, comes with limitations and normally requires a heavy downsampling of the event stream, thereby losing crucial information.

\subsection{Knowledge Distillation for Object Detection}
In recent years, knowledge distillation has been shown to be an effective approach for reducing the complexity of object detection models while maintaining high accuracy. 
It is usually classified into three types, response distillation~\cite{hinton2015distilling}, feature distillation~\cite{guo2021distilling, dai2021general, yang2022focal, yang2022masked, kang2021instance}, and label assignment distillation~\cite{nguyen2022improving}. 
Response distillation~\cite{hinton2015distilling, zheng2022localization} uses the final response distribution of the teacher model as soft labels for the student. 
Label assignment distillation~\cite{nguyen2022improving} leverages the teacher model to generate label assignments for the student.
Most distillation methods perform feature distillation, which works directly on feature maps. 
Defeat~\cite{guo2021distilling} highlights the importance of region selection for distillation.
ICD~\cite{kang2021instance} proposes a learnable conditional decoding module that uses each object instance as a query to distill the feature map with dot-product attention
FGD~\cite{yang2022focal} focuses on balancing foreground and background objects in spatial and channel-wise feature spaces. 
MGD~\cite{yang2022masked} utilizes a masked auto-encoder to align the feature space between the student and teacher. 

Different from classical knowledge distillation for object detection, there exists a significant domain gap between grayscale images and event frames, making it challenging to directly distill information between these two modalities. 
Intuitively, high-level knowledge, such as object information, has the potential to alleviate the modality gap between grayscale images and event frames.
To bridge the modality gap, recent studies have explored single-image-to-event translation techniques for feature representation~\cite{wang2021evdistill,messikommer2022bridging}, segmentation~\cite{sun2022ess}, and pre-training~\cite{yang2023event}. 
These studies have shown that task performance on events can be significantly improved by leveraging information from grayscale images. However, the use of direct image-to-event translation for object detection tasks has not been extensively investigated.
In contrast, cross-modality knowledge distillation has been utilized in other modalities to transfer knowledge, such as between RGB images and pseudo depth maps~\cite{chong2022monodistill}, and between RGB images and point clouds~\cite{hong2022cross,chen2022bevdistill}. Note that these methods have similar data requirements to ours as they need aligned cross-modal data during training but only use the student modality at inference. 

\begin{figure*}[htbp]
  \begin{center}
  \includegraphics[width=0.85\linewidth]{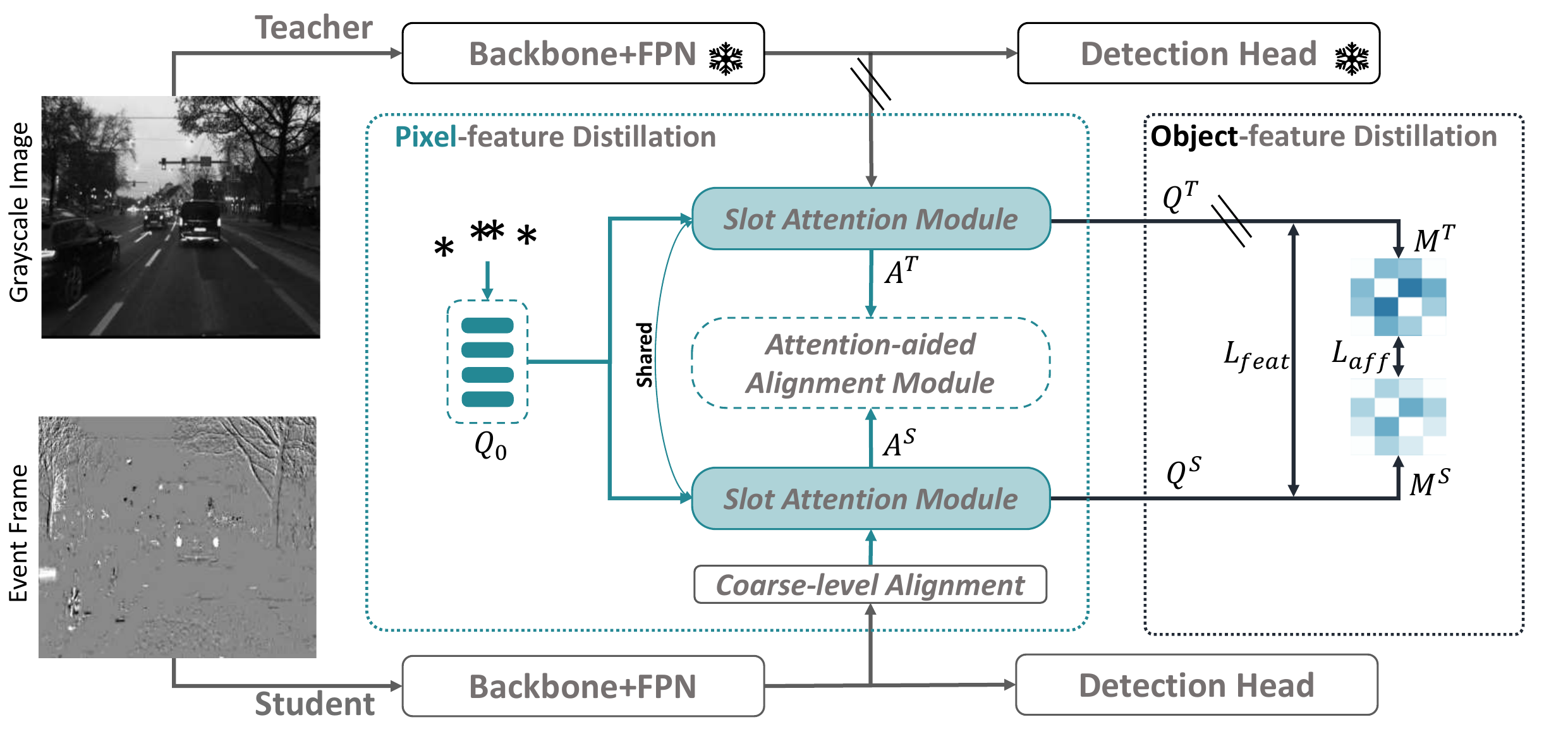}  
  \end{center}
  \vspace{-0.2cm}
  \caption{\textbf{Framework overview}. Our distillation model comprises two components: pixel-feature distillation and object feature distillation. The auxiliary task module is not drawn in the pixel-feature distillation.}
  \vspace{-0.9em}
  \label{fig:pipeline}
\end{figure*}
\section{Method}

\subsection{Overview}
Our cross-modality feature distillation pipeline consists of two main blocks, pixel-feature distillation, and object-feature distillation, which are illustrated in Fig.~\ref{fig:pipeline}. 
Specifically, our approach distills knowledge from the FPN output features, which enables us to apply the approach to both one-stage and two-stage detectors. 
Our method generates teacher features by passing grayscale images through a fully trained and frozen detection network. 
Meanwhile, our student model takes event voxel girds~\cite{gallego2020event} as input but otherwise uses an identical neural network. 
In the following sections, we will discuss each module in more detail and conclude with the combined loss function.
\begin{figure}[htbp]
  \begin{center}
  \includegraphics[width=0.8\linewidth]{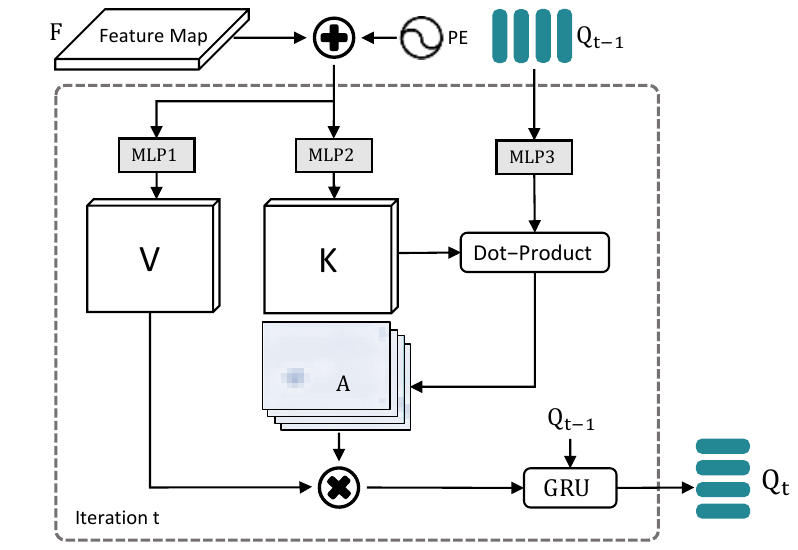}  
  \end{center}
  \vspace{-0.2cm}
  \caption{An illustration of the slot attention module in the t-th iteration. 
  %
  %
  }
  \vspace{-1.2em}
  \label{fig:slotattention}
\end{figure}

\subsection{Input Representations}\label{sec:input_rep}
Given the asynchronous event stream $e_i = \{(x_i, y_i, p_i, t_i) \}_{k=1}^N$, which consists of the $x$ and $y$ pixel locations, $\{-1, 1\}$ polarity and time step respectively. 
Our method takes a time window of duration $\Delta t$ and represents the event stream into a voxel grid representation following the approach in~\cite{zhu2019unsupervised}. 
The resulting voxel grid has a size of $B \times H \times W$, where $B$ is the number of temporal bins, and $H$ and $W$ are the height and width of the event frame, respectively. 
We also normalize the voxel grid values to the range $[0, 1]$ after clipping the pixel values to a fixed range to ensure a consistent input representation. 
The advantage of this event representation is that it can be efficiently implemented with a rolling buffer for real-time applications.
As for the teacher network input, we use grayscale images to avoid introducing further differences between the two modalities since events are also grayscale. 
\subsection{Object-centric Feature Extraction} 
To learn well-suited object-centric features, we use slot attention~\cite{locatello2020object}. 
The slot attention allows us to formulate an object-centric approach, with one slot feature per object.
This idea is closely related to our task, where we try to find features of object instances (distinct entities) that align well across modalities. 
Therefore, the slot attention module extracts a slot feature for each ground truth object, which is subsequently refined in each iteration, both for the student and the teacher model. 
The advantage of slot attention is that the location and feature of each object can be updated over the iterations, independently for the two modalities. 
Thus, it allows the method to find the best location and feature for knowledge distillation. 
Besides, from a practical point, this module also allows for some alignment errors between the two modalities.

As shown in Fig. \ref{fig:slotattention}, the input to the slot attention module is the FPN feature $F$, which is augmented with a positional embedding and then flattened into vectors. 
The resulting feature vector is denoted as $\overline{F}$. The second inputs are the $L$ initial query vectors, each containing $D$ features, resulting in $Q_0 \in \mathbb{R}^{D \times L}$. 
In the t-th step the query $q_{t-1} \in \mathbb{R}^{D \times L}$, key $K \in \mathbb{R}^{D \times HW}$, and value $V \in \mathbb{R}^{D \times HW}$ are generated via three separate linear transformations,

\vspace{-0.4cm}
\begin{equation}
    q_{t-1} = W^Q \cdot Q_{t-1}, \; K = W^K \cdot \overline{F}, \; V = W^V \cdot \overline{F}\,,
\end{equation}
where $W^Q$, $W^K$, $W^V \in \mathbb{R}^{D \times D}$.

Given the key, query, and value, we compute the output of the module using an attention update.
We normalize slot attention coefficients over the slot dimension~\cite{locatello2020object}.
This enforces a competition between slots to explain parts of the inputs while ensuring that each pixel is mainly assigned to one query. 
This also allows us to compute the attention matrix,
\vspace{-0.2cm}
\begin{equation}
    \text{attn}_{i,j} = \frac{e^{M_{i,j}}}{\sum_l e^{M_{i,l}}}, \; \text{with} \;
    M = \frac{1}{\sqrt{D}} q_{t-1}K^T\,,
\end{equation}
where $\text{attn} \in \mathbb{R}^{HW \times L}$. The input values and their assigned query vectors are then aggregated via a simple weighted mean by their co-attention,
\vspace{-0.2cm}
\begin{equation}
    U = V \cdot A, \; \text{with} \; A_{i,j} = \frac{\text{attn}_{i,j}}{\sum_l{\text{attn}_{l,j}}}\,,
\end{equation}
where $U \in \mathbb{R}^{D \times L}$. In the final step, the query slots $Q_{t-1}$ are updated with a Gated Recurrent Unit (GRU) to generate the slot query for the next step $Q_{t} = \text{GRU}(Q_{t-1}, U)$. The use of the GRU model allows us to smooth the updates and consider past queries. 
We use the feature from the object center location as the initial query feature to improve the efficiency of iterative updates.

\subsection{Pixel Feature Distillation}

\noindent\textbf{Coarse-level Feature Alignment.} 
To alleviate the modality gap between grayscale images and event frames, we first introduce a shallow network $\mathcal{G}_c$ that maps student features ${F}^\mathcal{S}$ to an intermediate latent space ${\widetilde{F}}^\mathcal{S}$. 
Mapping the student feature to the latent space forces the student model to adapt to the coarse feature distribution of the teacher.
All distillations are performed between the newly mapped student feature ${\widetilde{F}}^\mathcal{S}$ and the teacher feature ${F}^\mathcal{T}$.

\begin{figure}[htbp]
  \begin{center}
  \vspace{-0.2cm}
  \includegraphics[width=0.7\linewidth]{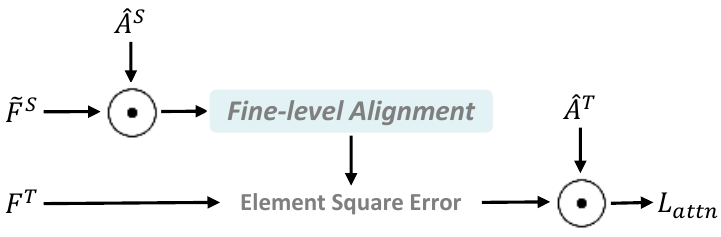}  
  \end{center}
  \vspace{-0.2cm}
  \caption{An illustration of the attention-aided fine-level feature alignment module. 
  %
  %
  }
  \vspace{-0.9em}
  \label{fig:pixelfeaturedistill}
\end{figure}
\noindent\textbf{Attention-aided Fine-level Feature Alignment.} 
During the slot attention module, the attention maps of the teacher and the student $\text{attn}^\mathcal{T}$ and $\text{attn}^\mathcal{S}$, learned to focus on the important region of the corresponding objects by interacting with the slot features. 
Thus, they contain crucial spatial information about the importance of the feature map for all objects. 
Technically, our attention-aided fine feature alignment module takes the latest attention maps and reshapes them into attention matrices for each object, $\mathcal{A}^\mathcal{T}_i$ and $\mathcal{A}^\mathcal{S}_i$, where the subscript $i$ refers to the $i$th object. 
These attention matrices are then used to mask the FPN feature maps ${\widetilde{F}}^\mathcal{S}$ and ${F}^\mathcal{T}$. 
As illustrated in Fig. \ref{fig:pixelfeaturedistill}, we first mask the student's feature map using the student attention map, 
\vspace{-0.1cm}
\begin{align}
    \Tilde{F}^\mathcal{S}_i = \hat{\mathcal{A}}^\mathcal{S}_i \odot {\widetilde{F}}^\mathcal{S}\,.
\end{align}
Where $\odot$ is the Hadamard product, and the output is a masked feature map per object. ``hat" denotes the gradient detach. 
Given the masked student feature, we pass them through a fine-level feature alignment module $\mathcal{G}_g$, which is similar to \cite{yang2022masked}. 
Given this refined masked student feature, we compute the feature alignment loss with the teacher feature map using element squared error $\ell_{ese}$. 
Thus, we can compute the attention-aided fine feature alignment loss as,
\vspace{-0.2cm}
\begin{align}
&L_{\text {attn}}= \sum_i^L\left({\ell_{ese}\left(\mathcal{G}_f(\tilde{F}^\mathcal{S}_i), F^\mathcal{T}\right) \odot \hat{A}^{\mathcal{T}}_i}\right) / \sum_i^L{\hat{A}^{\mathcal{T}}_i} \,,
\end{align}
where the element wise square error is masked with the teacher's attention map. 

\subsection{Object Feature Distillation}
\noindent\textbf{Direct Object-centric Feature Distillation.} The slot attention module generates a slot feature for each ground truth object for the teacher $Q_{T,i}^\mathcal{T}$ and the student $Q_{T,i}^\mathcal{S}$, where the subscript $i$ refers to the $i$th object. 
To explicitly encourage the student model to mimic the teacher at a high level, we formulate an $\ell_{1}$ loss between the student and teacher slot feature. 
Thus, the feature distillation loss can be written as 
\vspace{-0.1cm}
\begin{equation}
L_{\text {feat}}=\frac{1}{L}\sum_i^L{\ell_{1}\left({Q}^{\mathcal{S}}_{T,i}, \hat{Q}^{\mathcal{T}}_{T,i}\right)}\,,
\end{equation}
where $L$ is the number of ground truth objects and slot features, and the ``hat" $\hat{Q}$ indicates that the gradient is stopped, ensuring the student model has cleaner gradients.

\noindent\textbf{Object-centric Relation Distillation.} Besides the direct distillation of object-centric features, which we tackle using the slot features, also the feature relation between the objects contains information and we propose to use this relation to further improve the distillation.
Given that student features can never be perfectly aligned with teacher features, exploiting this secondary information of inter-object relations can further boost the distillation. 
Inspired by \cite{hou2022point}, we build an affinity matrix between object-centric features to embed their relationship. 
The affinity matrix score calculates the similarity of each pair of object-centric features. 
Specifically, each entry of the affinity matrix of the teacher $\mathbf{M}^\mathcal{T}$ can be computed as
\vspace{-0.2cm}
\begin{align}
    \mathbf{M}^\mathcal{T}(m, n)=&\frac{{{Q}_{m}^\mathcal{T}}^{\top} {{Q}_{n}^\mathcal{T}}}{\left\|{{Q}_{m}^\mathcal{T}}\right\|_2\left\|{{Q}_{n}^\mathcal{T}}\right\|_2}\,,
\end{align}
where $m,n$ denote the m-th and n-th object-centeric features, respectively. The affinity matrix of the student $\mathbf{M}^\mathcal{S}$ can be computed identically by using the student slot feature $Q^\mathcal{S}$. Given the two affinity matrices, we detach the gradient of the teacher and use a Mean Square Error (MSE) loss to force the student to mimic the inter-object relationship of the teacher,
\vspace{-0.2cm}
\begin{align}
L_{\text {aff}}=\ell_{mse}\left(\mathbf{M}^\mathcal{S}, \hat{\mathbf{M}}^\mathcal{T}\right)\,.
\end{align}

\noindent\textbf{Auxiliary Task.} The slot attention module allows us to generate both informative slot features and attention maps that are the core building blocks of our distillation approach. 
However, the slot attention is potentially under constraint, and the problem could learn a trivial solution. 
To avoid this issue, we introduce two auxiliary tasks to keep the teacher features informative~\cite{kang2021instance}. 
We use a classification task that is supervised using Binary Cross Entropy (BCE) and a bounding box size regression task that is supervised by an L1 loss, resulting in the following cost function,
\vspace{-0.15cm}
\begin{equation}
L_{\text{aux}}=\frac{1}{L}\sum_i^L{\ell_{bce}(C_i^{\mathcal{T}}, C_i^{gt})} + \frac{\beta}{L} \sum_i^L{\ell_{1}(B_i^{\mathcal{T}}, B_i^{gt})}\,,
\end{equation}
where $C^{gt}$ and $B^{gt}$ denotes ground-truth classification and bounding box center location. $C_i^{\mathcal{T}}$ and $B_i^{\mathcal{T}}$ are the outputs of the auxiliary MLP $\mathcal{G}_\text{aux}({Q}^{\mathcal{T}})$, and $\beta$ is a hyper-parameter to balance the auxiliary task.

\subsection{Overall Loss}
The total loss of the student model can be written as
\vspace{-0.15cm}
\begin{equation}
L_{\text {total}}=L_{\text {det}} + \lambda_1L_{\text {feat}} + \lambda_2L_{\text {attn}} + \lambda_3L_{\text {aff}} + \lambda_4L_{\text {aux}}\, \label{eq:total_loss}
\end{equation}
where $L_{\text {det}}$ is the original detection loss of the student and $\{\lambda_1, \lambda_2, \lambda_3, \lambda_4\}$ are hyperparameters to balance the losses.
\section{Experiments}
\subsection{Experiment Setting}
We evaluate our method on two datasets, the first is a synthetic dataset generated using the CARLA Simulator~\cite{dosovitskiy2017carla}. The dataset is recorded in seven CARLA towns in good weather conditions, where one town is used as the validation sequence. The CARLA dataset has 29,600 labeled frames and contains three classes: vehicle, pedestrian, and two-wheeler. The second is the real-world event dataset DSEC~\cite{gehrig2021dsec}. We generate bounding boxes by annotating the RGB images using Cascade-Faster R-CNN~\cite{cai2018cascade} trained on BDD100k~\cite{yu2020bdd100k}. The RGB detection results are then projected to the event camera using a mean depth assumption, which works well due to the small baseline. Similarly, to align images between the two modalities, we warp the RGB images to the event camera given a mean depth. Note that we only auto-label the DSEC training split, which consists of 53 driving sequences, and use a subset of 5 sequences as a validation set. The resulting DSEC dataset has 52,645 labeled frames and contains seven classes: pedestrian, rider, car, truck, bus, bicycle, and motorcycle.

Note that for other real-world datasets, a similar approach can be used to align the RGB and event data, as long as the event and RGB cameras are mounted close to each other. This dataset requirement is fulfilled for a large number of existing event datasets~\cite{gehrig2021dsec, zhu2018multivehicle, Chaney_2023_CVPR, tulyakov2021time,mitrokhin2019ev}, which contain both RGB and event data. 

As discussed in Section~\ref{sec:input_rep}, we use a normalized voxel gird as our input representation, in the case of CARLA we use a time window of $\Delta t=50$ms with $B=5$ temporal bins. For DSEC we use a larger temporal window of $\Delta t=100$ms with $B=10$ temporal bins, thus in both cases each temporal bin contains $10$ms of event data.

For our main experiment, we use YOLOX~\cite{ge2021yolox} a fast one-stage anchor-free detector, since it well represents modern real-time detectors~\cite{redmon2018yolov3, bochkovskiy2020yolov4, ge2021yolox, li2022yolov6}. Furthermore, to show that our distillation approach generalizes to other detector architectures, we use Faster R-CNN \cite{girshick2015fast} which is an anchor-based two-stage detector. For YOLOX, we use the ``small" version, which uses 9M parameters and achieves run times of below 10ms in PyTorch eager mode and less than 2ms using TensorRT on an Nvidia 3090. This allows for real-time inference using a rolling buffer in our event representation with time bins of 10ms. For Faster R-CNN we use a ResNet-50~\cite{he2016deep} as backbone. We use $T=3$ iterations in the slot attention module, and hyperparameters in Eq.~\ref{eq:total_loss} are set to $\lambda=\{0.1, 1.0, 0.002, 1.0\}, \beta=50$ for YOLOX and $\lambda=\{0.01, 0.1, 0.0002, 0.1\}, \beta=50$ for Faster R-CNN.

\definecolor{LightCyan}{rgb}{0.88,1,1}
\definecolor{Gray}{gray}{0.85}
\begin{table}[!t]
\centering
\caption{\label{tab:comparison_carla} Comparison with SOTA methods on the CARLA val set. $\dagger$ denotes without response distillation.}
\vspace{-0.2cm}
\scalebox{1.0}{
\begin{tabular}{l|lll}
\hline
Method & mAP & AP$_{50}$ & AP$_{75}$\\ \hline
YOLOX-S (Teacher-GS) &59.6 &81.1 &70.8\\
YOLOX-S (Student-Event) &53.8 &72.6 &58.4\\ \hline
+ ICD\cite{kang2021instance} &53.1 (-0.7) &73.7 &60.5 \\
+ FGD\cite{yang2022focal} &53.8 (+0.0) &75.9 &61.2\\
+ MGD\cite{yang2022masked} &54.7 (+0.9) &73.2 &60.8\\
+ MonoDistill$\dagger$\cite{chong2022monodistill} &54.1 (+0.3) &72.5 &61.7\\ 
\rowcolor{Gray} + Ours  &56.4 (+2.6) &76.4  &63.2\\\hline 
Faster R-CNN (Teacher-GS) &49.6 &71.6 &57.5\\
Faster R-CNN (Student-Event) &41.8 &63.4 &45.7\\ \hline
+ ICD\cite{kang2021instance} &41.6 (-0.2) &62.0 &46.5\\
+ FGD\cite{yang2022focal} &42.7 (+0.9) &64.0 &46.4\\
+ MGD\cite{yang2022masked} &40.1 (-1.7) &60.7 &44.3\\
+ MonoDistill$\dagger$\cite{chong2022monodistill} &38.6 (-3.2) &58.8 &43.2\\ 
\rowcolor{Gray} + Ours &44.0 (+2.2) &63.4 &49.5\\\hline 
\end{tabular}
}
\vspace{-2.0em}
\end{table}
\subsection{Main Results}
We compare our novel cross-modality distillation method with four state-of-the-art distillation methods for object detection, including three single modality distillation methods, ICD~\cite{kang2021instance}, FGD~\cite{yang2022focal}, MGD~\cite{yang2022masked}, and one cross-modality distillation method, MonoDistill~\cite{chong2022monodistill}. For all methods, we perform feature distillation on the multi-scale FPN features. 

The results of our comparison experiment on the CARLA dataset are presented in Table \ref{tab:comparison_carla}. The majority of the state-of-the-art methods result in a decrease in the performance of the student network. 
However, our distillation method significantly outperforms the other four methods. The student detector with our distillation module demonstrates significant mAP improvement. Our method results in 2.6 mAP and 2.2 mAP improvements on YOLOX and Faster R-CNN, respectively. When comparing this to the grayscale teacher model, our distillation method is able to nearly half the performance gap to the teacher model.

\definecolor{LightCyan}{rgb}{0.88,1,1}
\definecolor{Gray}{gray}{0.85}
\begin{table}[htb]
\centering
\caption{\label{tab:comparison_dsec_ema} Comparison with SOTA methods on the DSEC val dataset. All student models are updated by Exponential Moving Average.}
\vspace{-0.2cm}
\scalebox{1.0}{
\begin{tabular}{l|lll}
\hline
Method & mAP & AP$_{50}$ & AP$_{75}$\\ \hline
YOLOX-S (Teacher-GS) &47.5 &65.7 &53.3\\
YOLOX-S (Student-Event) &27.4 &42.9 &29.0   \\ \hline
+ ICD &28.1 (+0.7) &44.6 &29.9\\
+ FGD &29.8 (+2.4) &46.5 &31.9\\
+ MGD &28.8 (+1.4) &45.9 &30.8\\
+ MonoDistill$\dagger$ &29.4 (+2.0) &46.3 &30.7\\
\rowcolor{Gray} + Ours &30.8 (+3.4) &49.3 &32.4 \\\hline 
\end{tabular}
}
\vspace{-0.8em}
\end{table}
In Table~\ref{tab:comparison_dsec_ema} we compare our distillation method on the DSEC dataset. In the case of DSEC, the grayscale teacher model is clearly better than the event based detector. We attribute this to the auto-labeling, which potentially misses objects that are hard to detect in the RGB images, giving frame based detectors an advantage. When investigating state-of-art-methods, we can see that distillation is definitely beneficial for event based detectors, with all methods improving the detection performance. However, our method results clearly the largest improvement with an increase of 3.4 mAP.

\subsection{Ablation Studies and Analysis}
To evaluate the individual components of our proposed method, we conduct a series of ablation experiments using the YOLOX detector on the CARLA dataset.

\vspace{-0.2cm}
\begin{table}[htbp]
\centering
\caption{\label{tab:ablation_component} Ablation on feature distillation components. 
}
\vspace{-0.2cm}
\scalebox{0.95}{
\begin{tabular}{cccc|llllll}
\hline
Coarse & Fine+Direct & Relation  & Auxiliary  & mAP  & AP$_{50}$  & AP$_{75}$\\ \hline
\XSolidBrush  & \XSolidBrush  & \XSolidBrush  & \XSolidBrush&53.8  &72.6  &58.4\\ \hline 
\CheckmarkBold  & \CheckmarkBold  & \XSolidBrush  & \CheckmarkBold  &55.3  &77.1  &63.0\\ 
\XSolidBrush  & \CheckmarkBold  & \XSolidBrush & \CheckmarkBold  &52.0  &73.1  &61.9\\
\XSolidBrush  &  \CheckmarkBold  &  \XSolidBrush  & \XSolidBrush  &37.9  &64.7  &39.7\\\hline
\rowcolor{Gray} \CheckmarkBold  &    \CheckmarkBold  &  \CheckmarkBold  & \CheckmarkBold  &56.4  &76.4  &63.2\\\hline 
\end{tabular}
}
\vspace{-1.4em}
\end{table}

\noindent\textbf{Module Component.}
To gain a better understanding of our model component, we conduct ablation studies on the various components. Starting with a baseline mAP of 53.8 without distillation, we add a base module comprising of coarse-level feature alignment, attention-aided fine-level feature alignment, direct object-centric feature distillation, and auxiliary task, resulting in a 1.5 mAP improvement. Removing the coarse-level feature alignment module from the basic module causes a significant 3.3 mAP drop in performance, demonstrating the crucial role of this module in aligning event and grayscale features. Furthermore, removing the auxiliary task causes the model's performance to drop drastically to 37.9 mAP, indicating that without the auxiliary task, our object-centric feature extraction module learns a shortcut resulting in a trivial problem. Finally, adding object-centric relation distillation to our basic module improves performance by 1.1 mAP, showing that incorporating relational information between instance features can help bridge the modality gap and enhance performance.

\begin{table}[htpb]
\centering
\caption{\label{tab:ablation_attention} Ablation on attention types.}
\vspace{-0.2cm}
\scalebox{1.0}{
\begin{tabular}{c|lll}
\hline
Attention Type & mAP           &AP$_{50}$ &AP$_{75}$\\ \hline
 -             &53.8           &72.6    &58.4\\ \hline
Full Region    &54.8 (+1.0)    &76.5    &62.0\\
FG Region      &55.0 (+1.2)    &77.0    &60.6\\\hline
FGD Full       &53.8 (+0.0)    &75.9    &61.2\\
FGD FG         &54.6 (+0.8)    &72.8    &60.8\\\hline
\rowcolor{Gray} Ours &55.3 (+1.5)&77.1  &63.0\\\hline 
\end{tabular}
}
\vspace{-0.4em}
\end{table}
\noindent\textbf{Attention Type.}\label{sec:attention_type}
To investigate the different attention types, we test several alternatives to our method that use the attention matrix of the slot attention module, see Table~\ref{tab:ablation_attention}. ``Full Region" is the simplest approach, which distills the entire feature map only using a coarse alignment network. ``FG Region" improves the previous approach by only distilling the Foreground (FG) regions using the ground truth bounding boxes as masks. We also employ an advanced distillation method based on FGD~\cite{yang2022focal}. Similar to the two previous methods, we use original FGD over the full feature maps ``FGF Full" and over the foreground regions ``FGD FG". Ours is the one without object-centric relation distillation.

We can see that focusing on the foreground regions generally performs better independent of the method, validating our argument that foreground features align better between the two modalities. We argue that this is the case because foreground objects contain more salient and discriminating features compared to background regions, and hence, foreground regions should be given more attention during feature distillation. Conversely, background regions may introduce noise and affect the distillation performance adversely. 

\begin{figure}
  \begin{center}
  \includegraphics[width=0.62\linewidth]{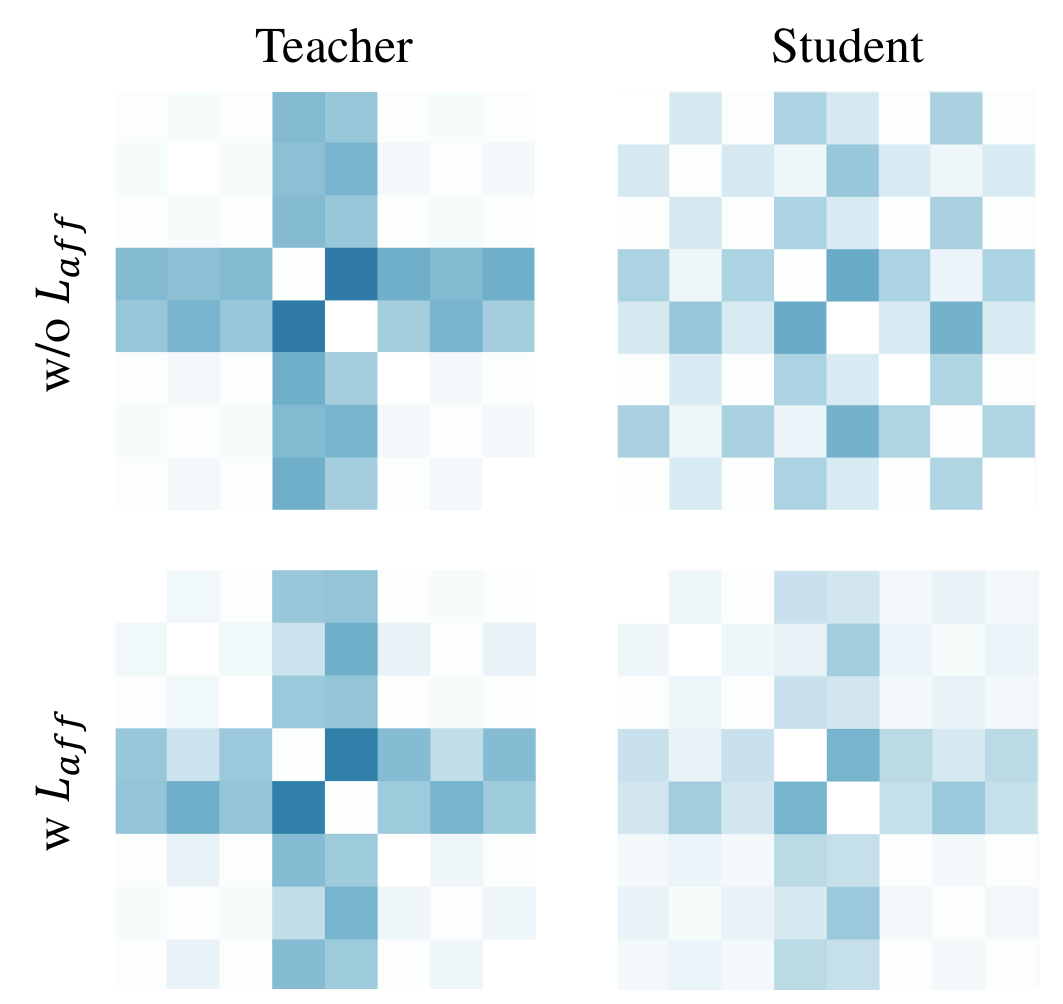}  
  \end{center}
  \vspace{-0.2cm}
  \caption{Visualization of the affinity matrix without and with object-centric relation loss. Affinity values are cosine similarity scores between object-centric features.}
  \label{fig:vis_aff}
  \vspace{-2.0em}
\end{figure}
\noindent\textbf{Object-centric Relation Distillation.} To better understand our object-centric relation distillation, we visualize the affinity matrix between slot features. In Fig.~\ref{fig:vis_aff}, we see that the affinity matrix generated by the teacher model exhibits high discrimination, while the affinity matrix from the student model appears to be sparse and noisy. With the integration of our relation distillation module, the student model produces a more structured affinity matrix that closely corresponds to the teacher. This indicates that our affinity distillation module enables a better transfer of structural knowledge from teacher to student. 

\noindent\textbf{Teacher Modality.} To validate our decision to use grayscale images as the teacher modality, we also perform an ablation using RGB images. Using RGB images improves the teacher to 62.7 mAP on CARLA compared to 59.6. However, the distillation does not work well and the performance is even false below not using distillation to 51.1 mAP.

\definecolor{LightCyan}{rgb}{0.88,1,1}
\definecolor{Gray}{gray}{0.85}
\begin{table}[htbp]
\caption{\label{tab:comparison_dsec_mc} Model compression task on the DSEC dataset.}
\vspace{-0.2cm}
\centering
\scalebox{1.0}{
\begin{tabular}{l|l|lll}
\hline
Method & YOLOX & mAP & AP$_{50}$ &AP$_{75}$\\ \hline
Teacher(GS) &S &47.5 &65.7 &53.3\\
Student(GS) &Tiny &43.5 &61.0 &50.1\\ \hline
+ ICD &Tiny &45.7 (+2.2) &65.7 &51.3    \\
+ FGD &Tiny &45.8 (+2.3) &64.7 &53.3\\
+ MGD &Tiny &45.4 (+1.9)&    64.5&    50.8\\
+ MonoDistill$\dagger$ &Tiny &45.9 (+2.4) &65.1 &51.6\\
\rowcolor{Gray} + Ours &Tiny &46.0 (+2.5) &64.4 &51.8\\
\hline 
\end{tabular}
} 
\vspace{-1.2em}
\end{table}
\subsection{Generalization Abilities}\label{sec:generalization}
Although our method is not designed for model compression, it is still interesting to investigate the detection performance within a single modality. 
Surprisingly, as demonstrated in Table~\ref{tab:comparison_dsec_mc}, our method performs competitively, with the best mAP improvement, showing that the method has also the potential for compression.

\section{Conclusion}
In this paper, we propose a cross-modality object detection distillation method based on slot attention to minimize the modality gap between events and images in object detection. We show that distilling object-centric features with corresponding attention maps, rather than directly mimicking image features can significantly improve event based detection performance. We conduct extensive experiments on synthetic and real event datasets to prove the effectiveness of our approach. 

{\small
\bibliographystyle{ieee_fullname}
\bibliography{egbib}
}

\section*{APPENDIX}
\begin{figure*}[htbp]
\centering
\begin{tabular}{cccccc}
\rotatebox{90}{Grayscale Input} &\includegraphics[width=0.17\textwidth]{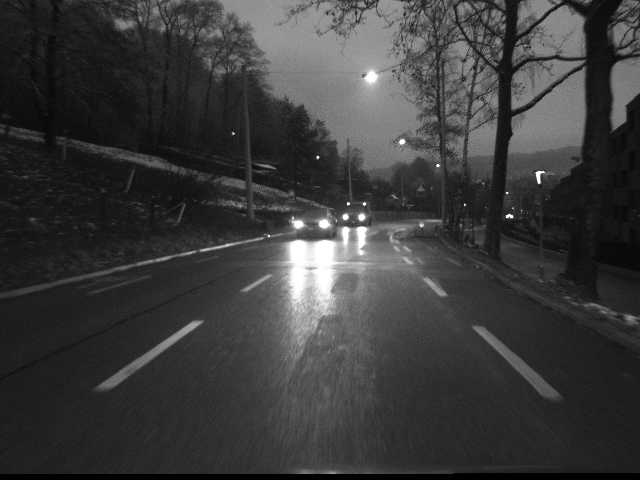}&
\includegraphics[width=0.17\textwidth]{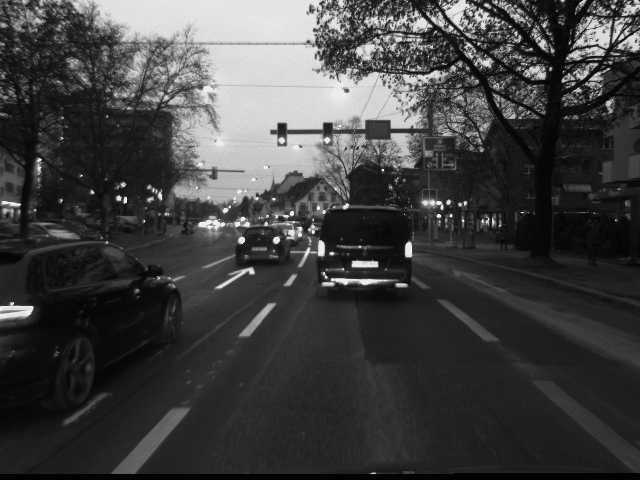}&
\includegraphics[width=0.17\textwidth]{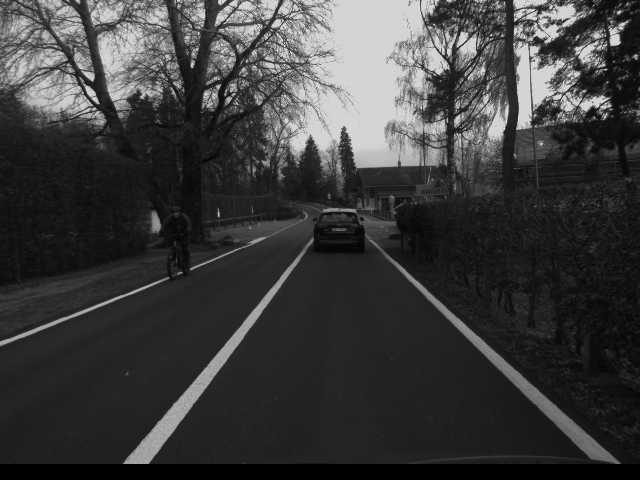}&
\includegraphics[width=0.17\textwidth]{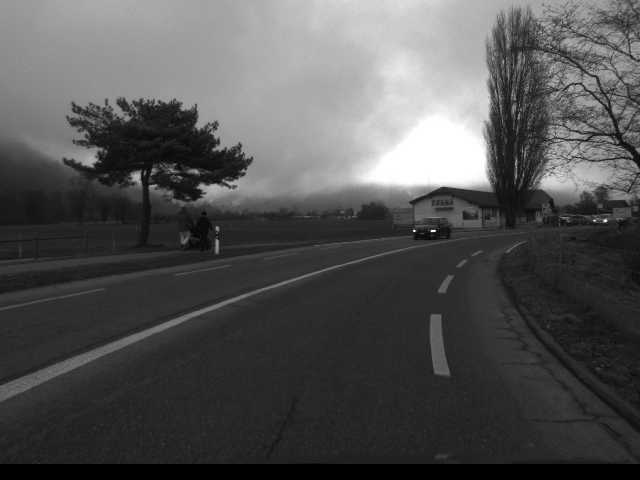}&
\includegraphics[width=0.17\textwidth]{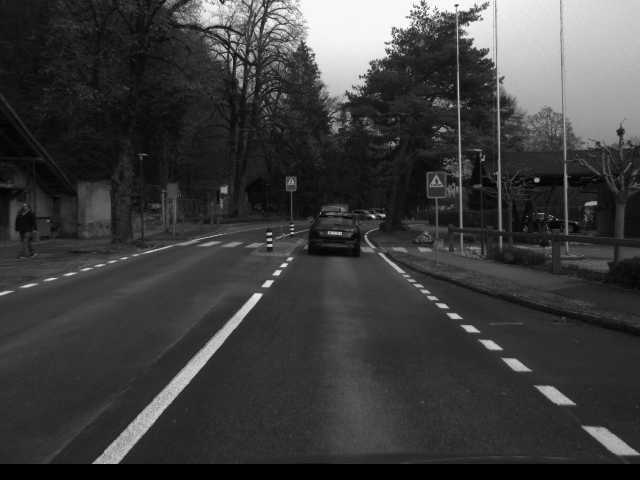}\\

\rotatebox{90}{~Events Input} &\includegraphics[width=0.17\textwidth]{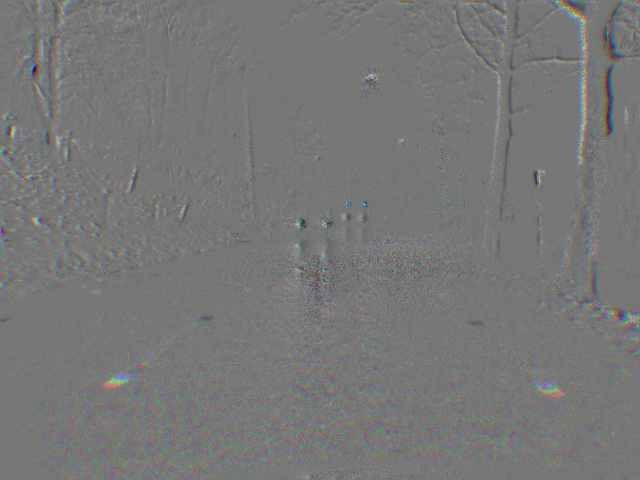}&
\includegraphics[width=0.17\textwidth]{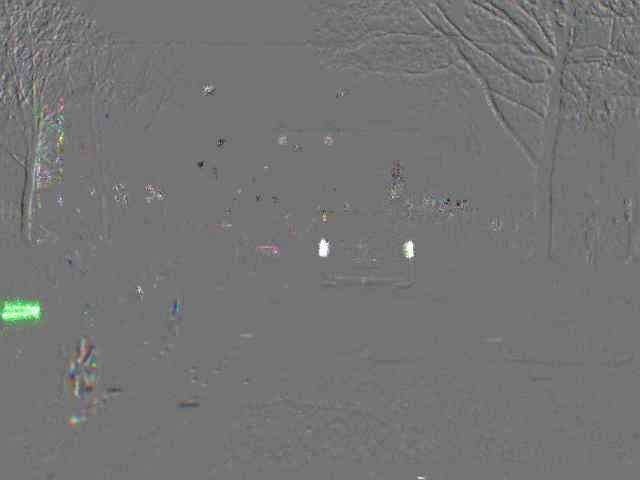}&
\includegraphics[width=0.17\textwidth]{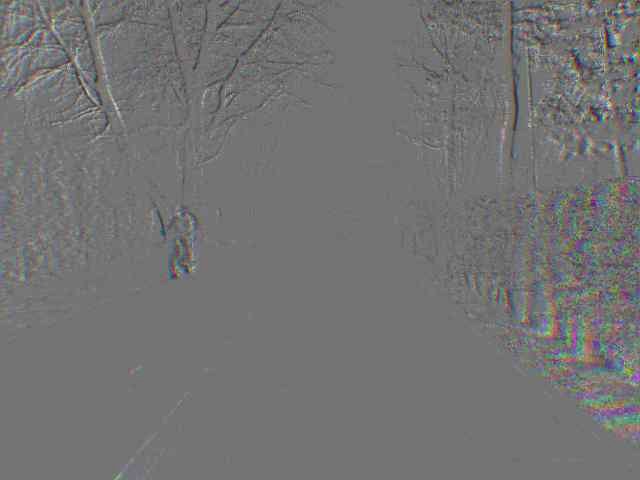}&
\includegraphics[width=0.17\textwidth]{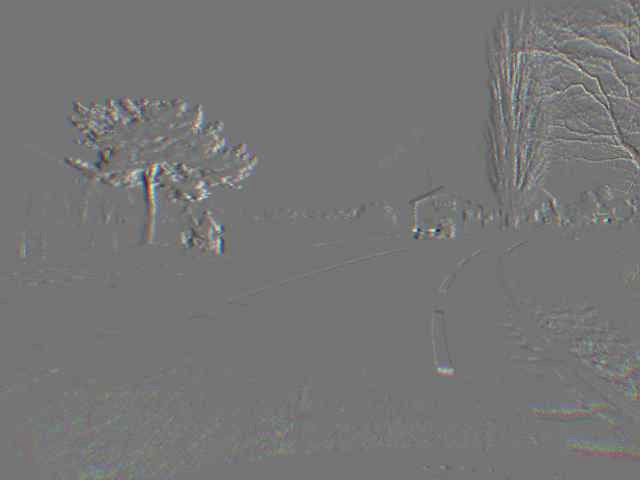}&
\includegraphics[width=0.17\textwidth]{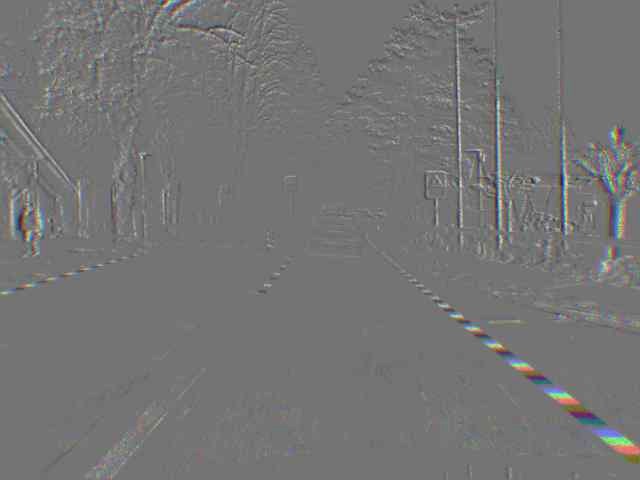}\\

\rotatebox{90}{~~~~~Grayscale} &\includegraphics[width=0.17\textwidth]{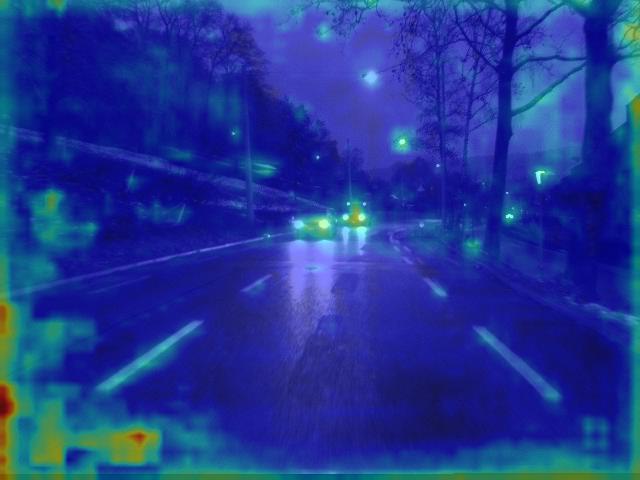}&
\includegraphics[width=0.17\textwidth]{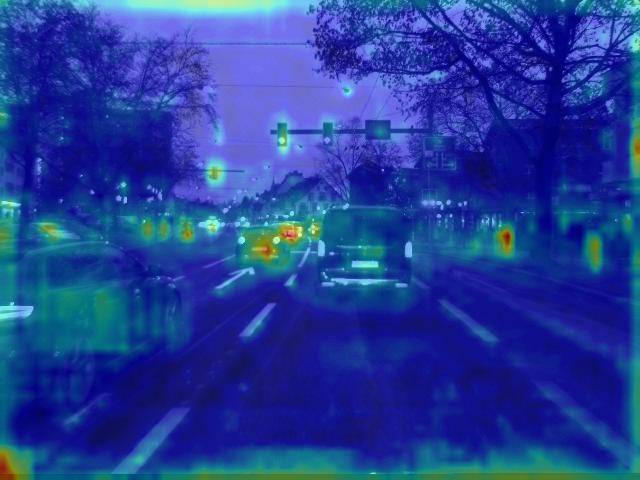}&
\includegraphics[width=0.17\textwidth]{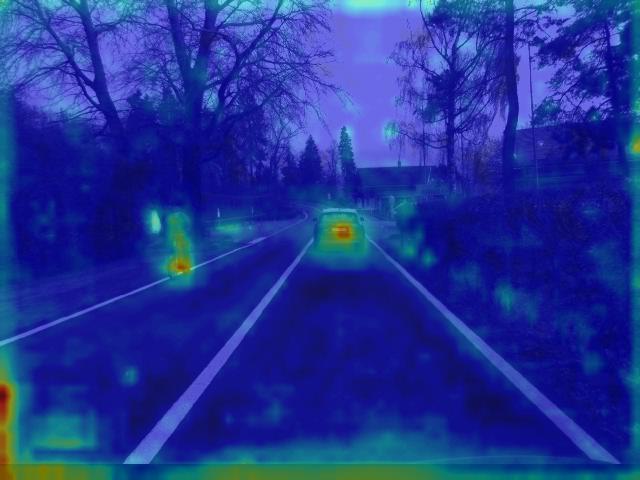}&
\includegraphics[width=0.17\textwidth]{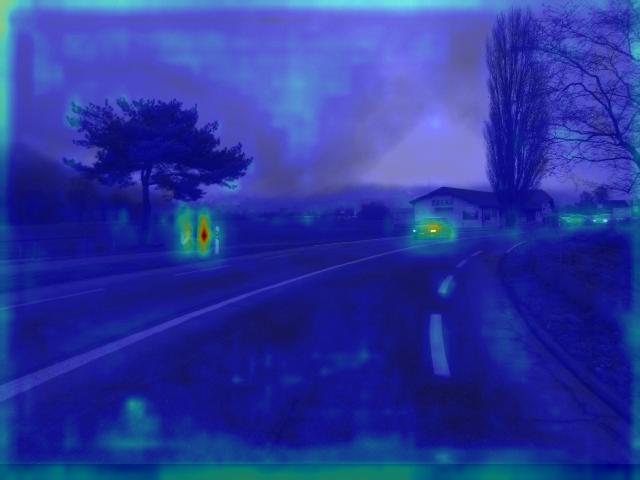}&
\includegraphics[width=0.17\textwidth]{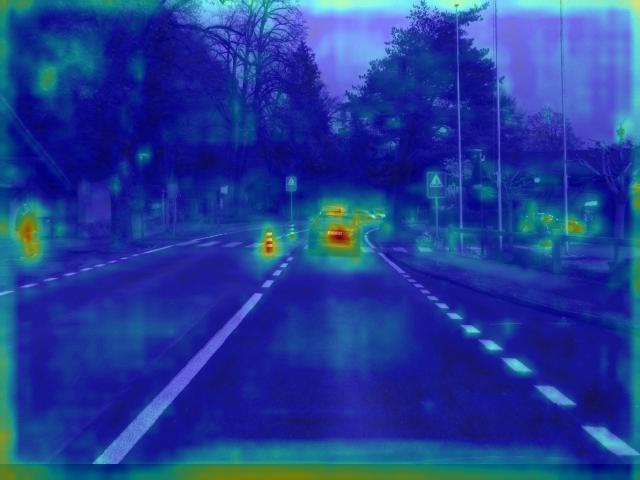}\\

\rotatebox{90}{~~~~~~~~Events} &\includegraphics[width=0.17\textwidth]{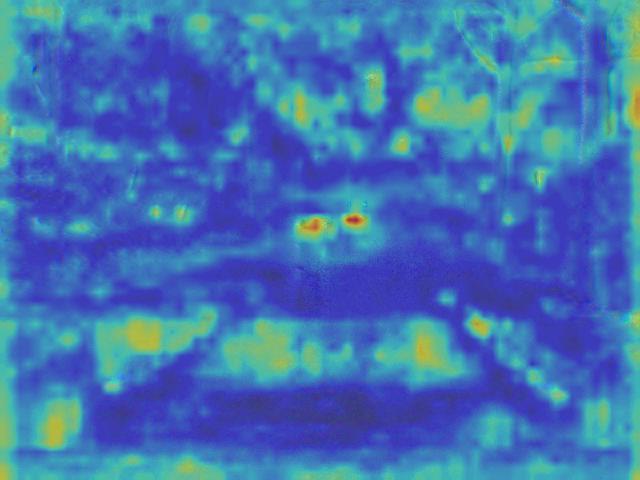}&
\includegraphics[width=0.17\textwidth]{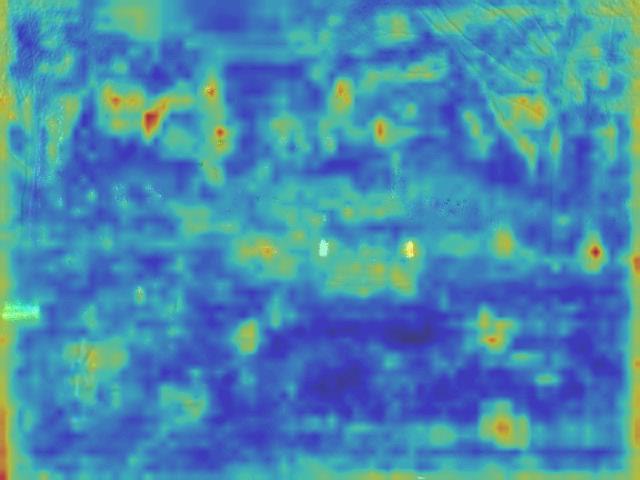}&
\includegraphics[width=0.17\textwidth]{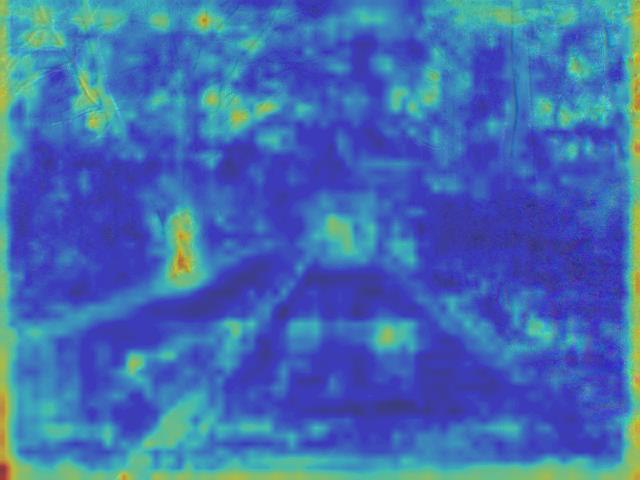}&
\includegraphics[width=0.17\textwidth]{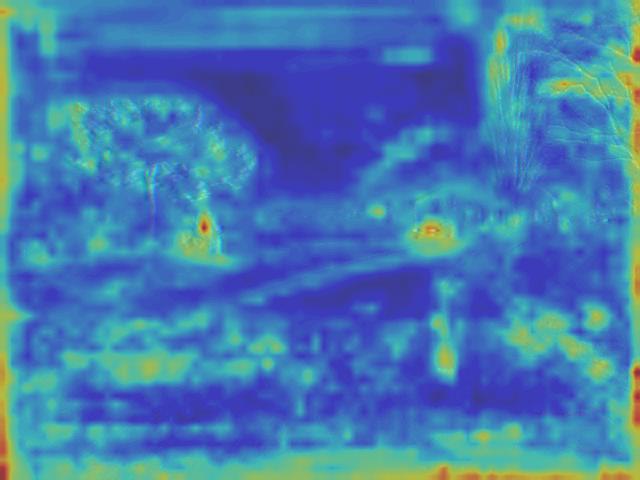}&
\includegraphics[width=0.17\textwidth]{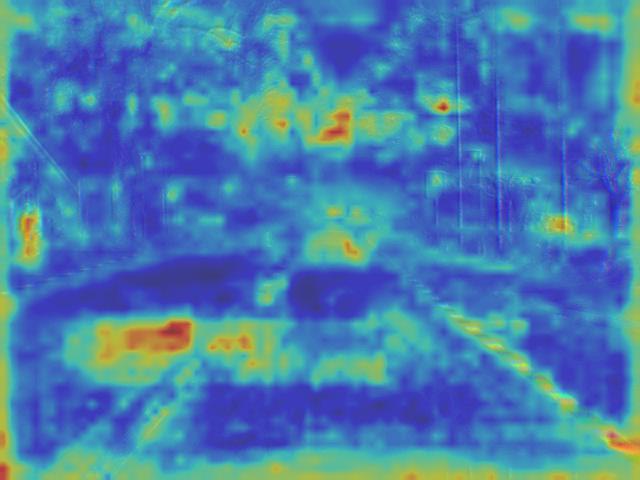}\\

\rotatebox{90}{~~~~~~~~+ICD} &\includegraphics[width=0.17\textwidth]{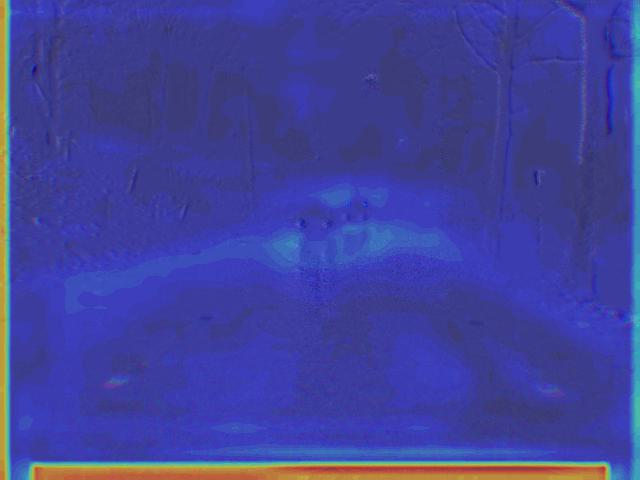}&
\includegraphics[width=0.17\textwidth]{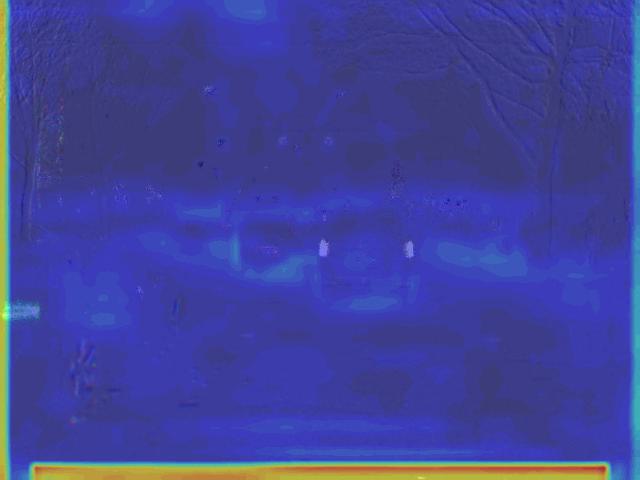}&
\includegraphics[width=0.17\textwidth]{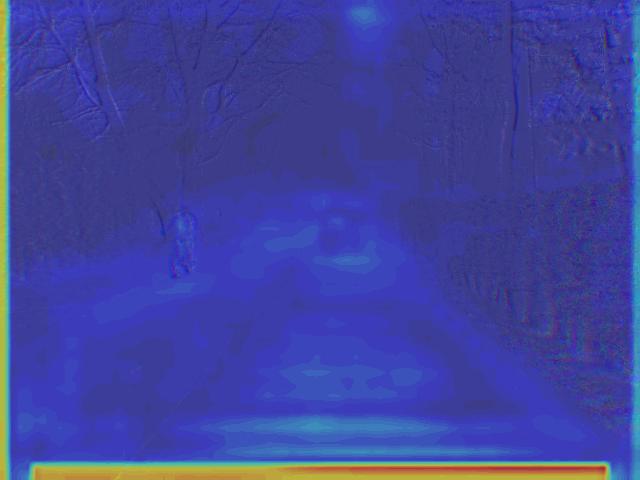}&
\includegraphics[width=0.17\textwidth]{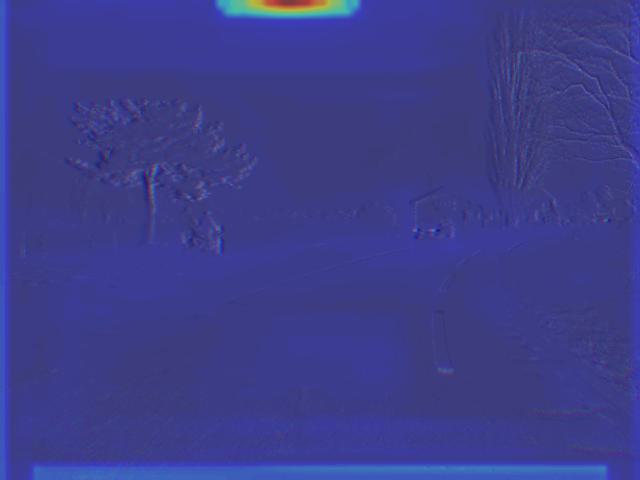}&
\includegraphics[width=0.17\textwidth]{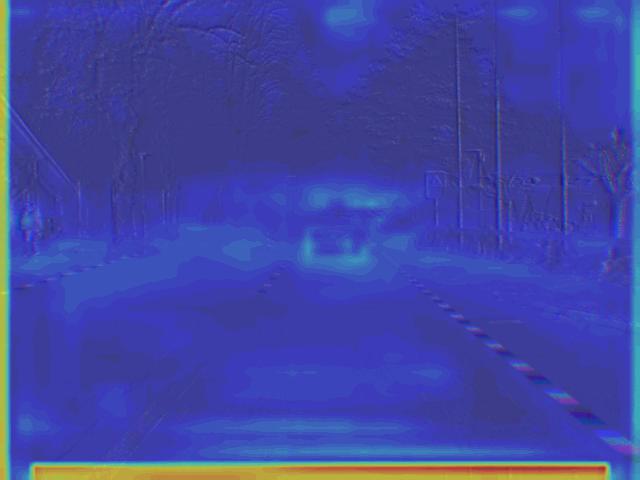}\\

\rotatebox{90}{~~~~~~~~+FGD} &\includegraphics[width=0.17\textwidth]{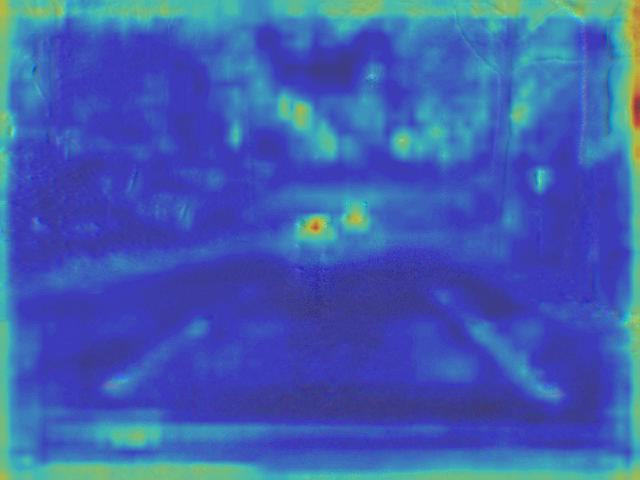}&
\includegraphics[width=0.17\textwidth]{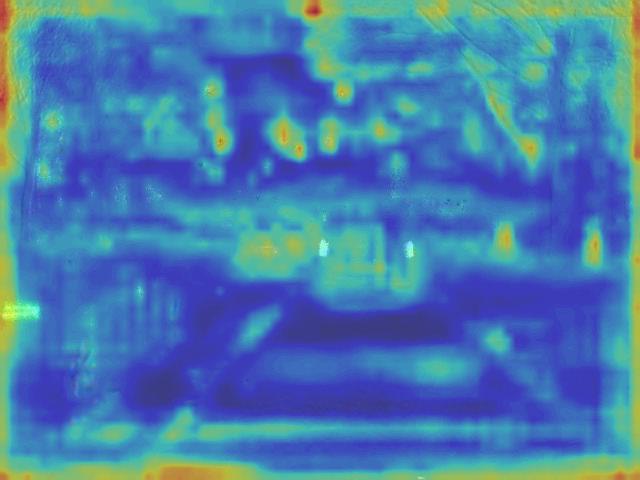}&
\includegraphics[width=0.17\textwidth]{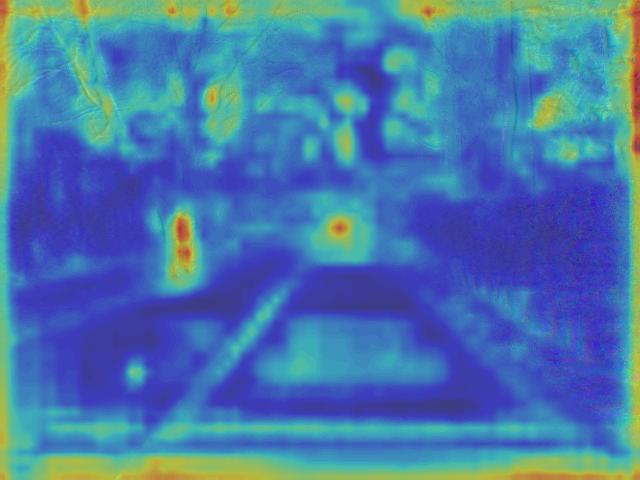}&
\includegraphics[width=0.17\textwidth]{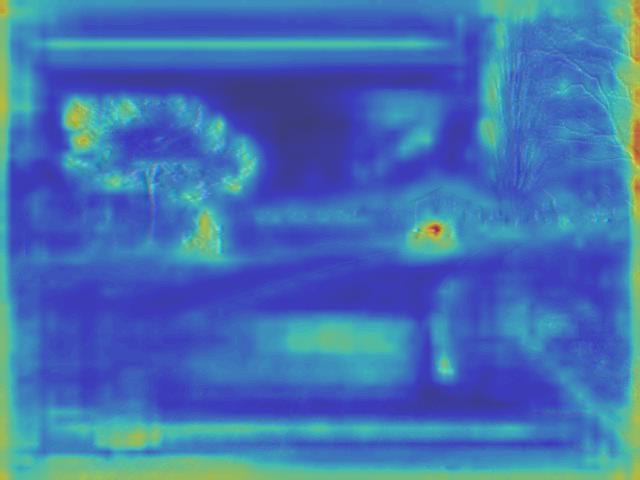}&
\includegraphics[width=0.17\textwidth]{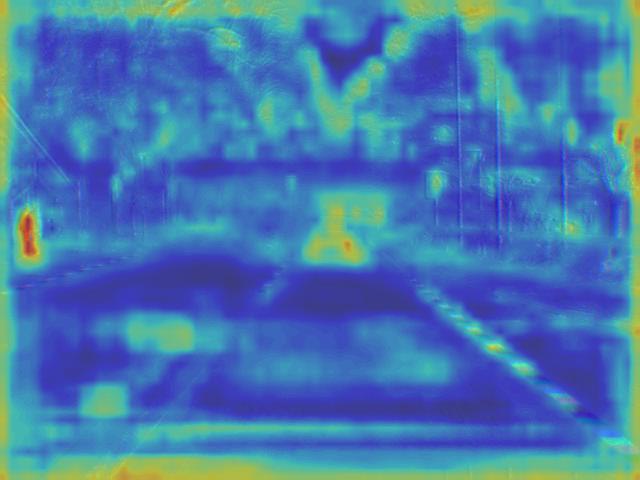}\\

\rotatebox{90}{~~~~~~~~+MGD} &\includegraphics[width=0.17\textwidth]{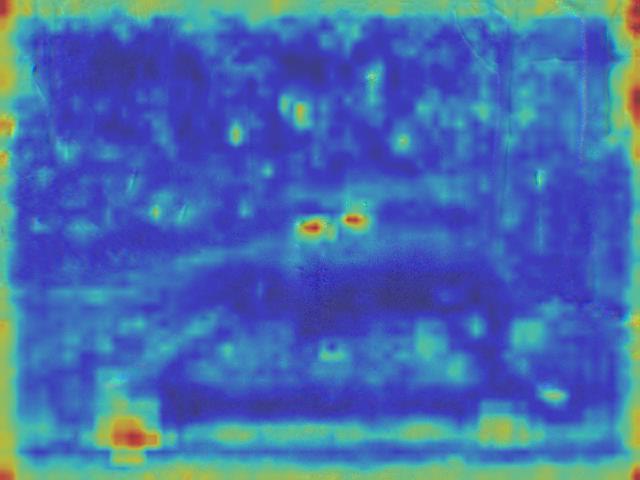}&
\includegraphics[width=0.17\textwidth]{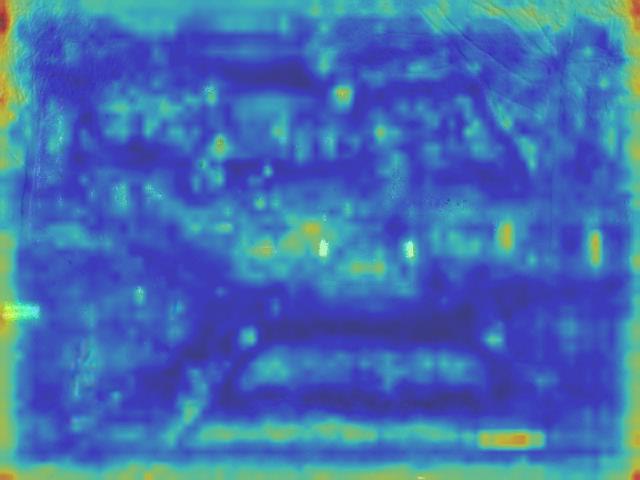}&
\includegraphics[width=0.17\textwidth]{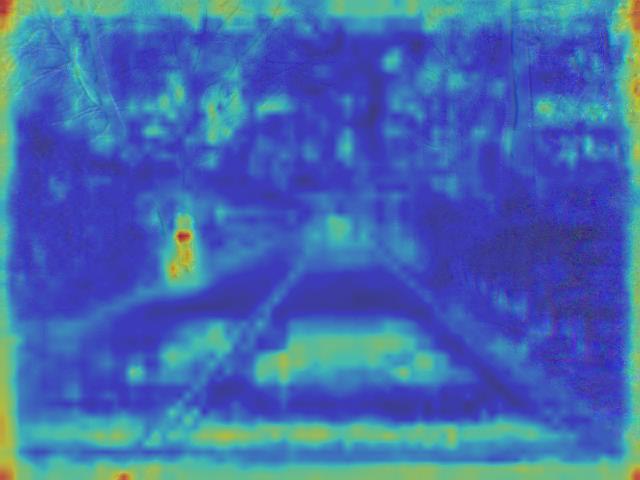}&
\includegraphics[width=0.17\textwidth]{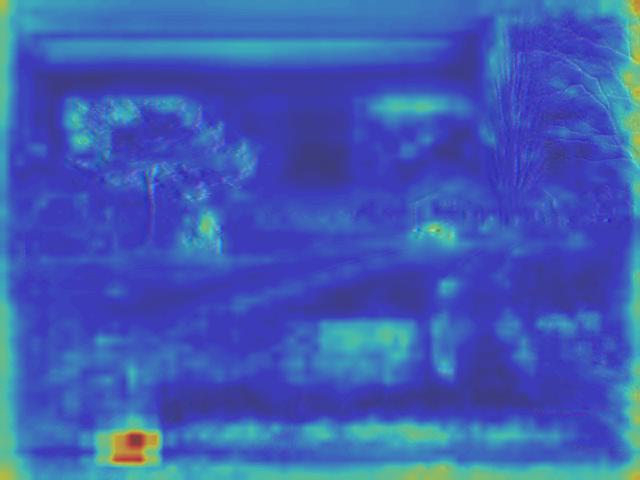}&
\includegraphics[width=0.17\textwidth]{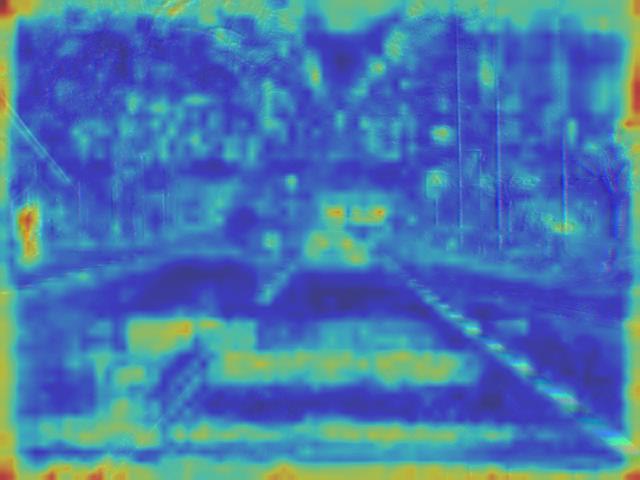}\\

\rotatebox{90}{~+MonoDistill} &\includegraphics[width=0.17\textwidth]{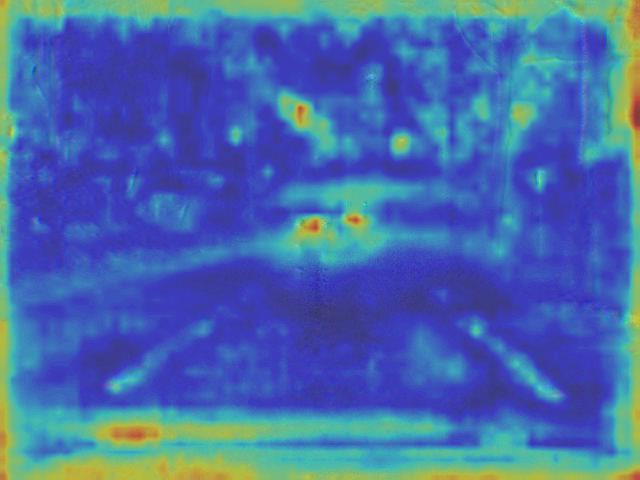}&
\includegraphics[width=0.17\textwidth]{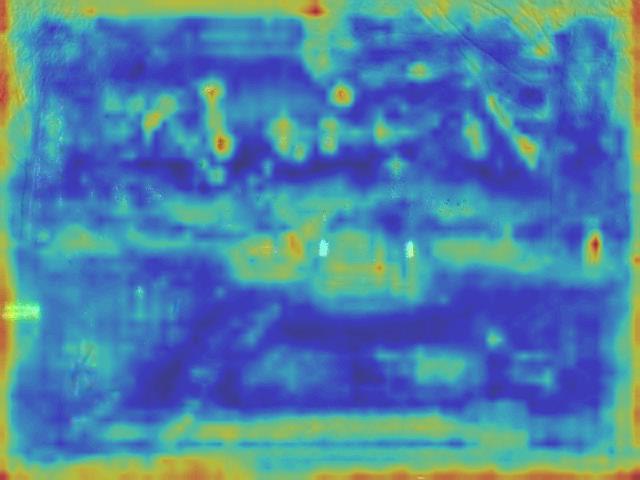}&
\includegraphics[width=0.17\textwidth]{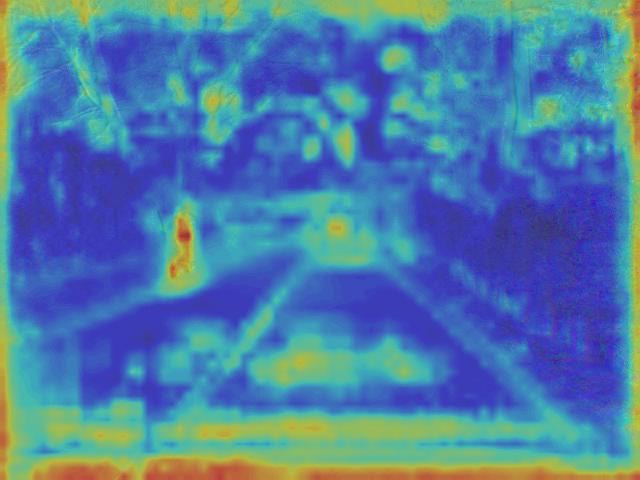}&
\includegraphics[width=0.17\textwidth]{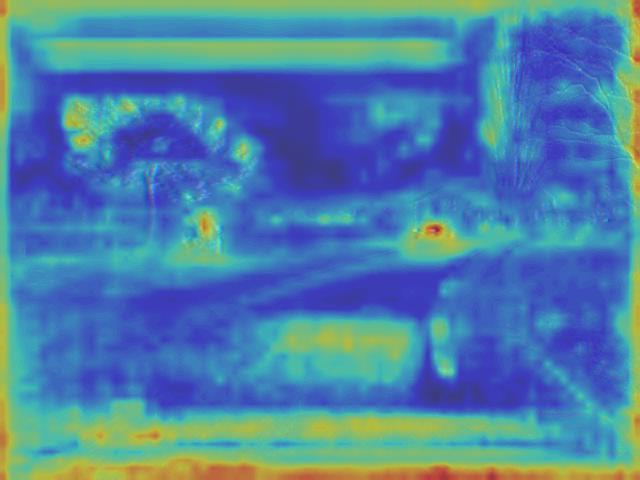}&
\includegraphics[width=0.17\textwidth]{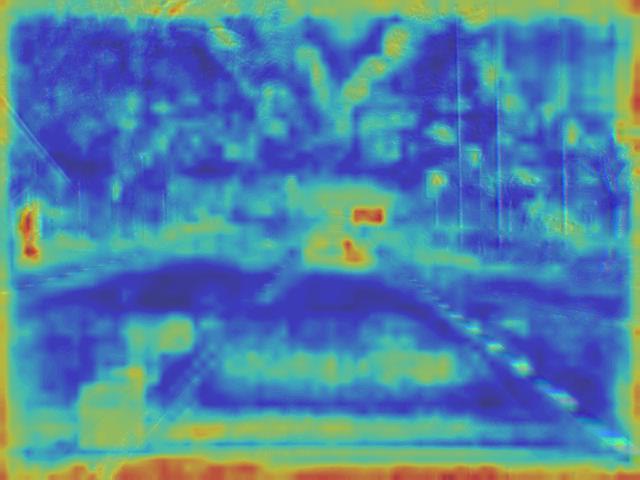}\\

\rotatebox{90}{~~~~~~~~~+Ours} &\includegraphics[width=0.17\textwidth]{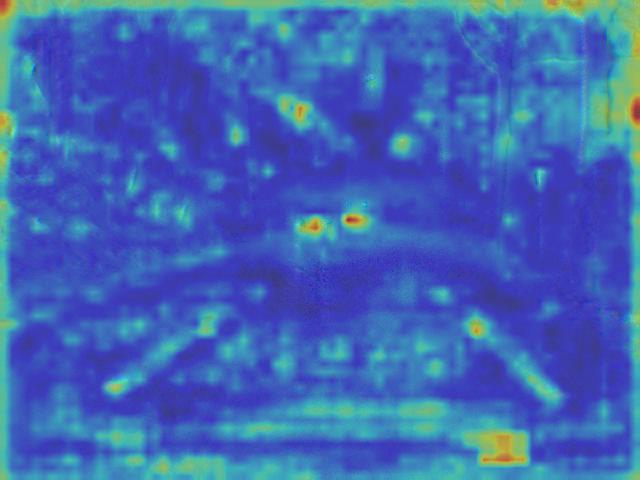}&
\includegraphics[width=0.17\textwidth]{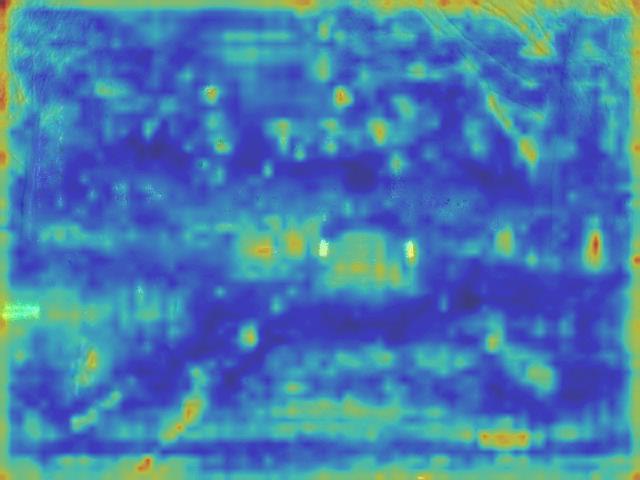}&
\includegraphics[width=0.17\textwidth]{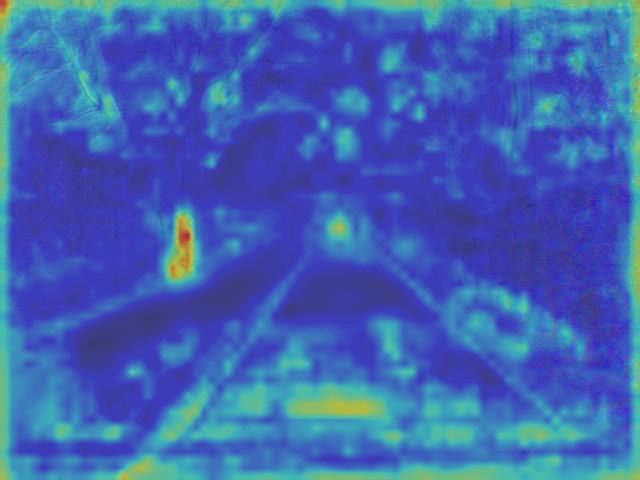}&
\includegraphics[width=0.17\textwidth]{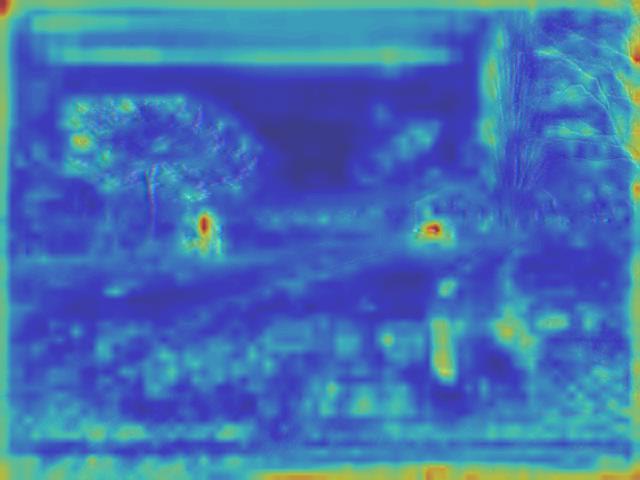}&
\includegraphics[width=0.17\textwidth]{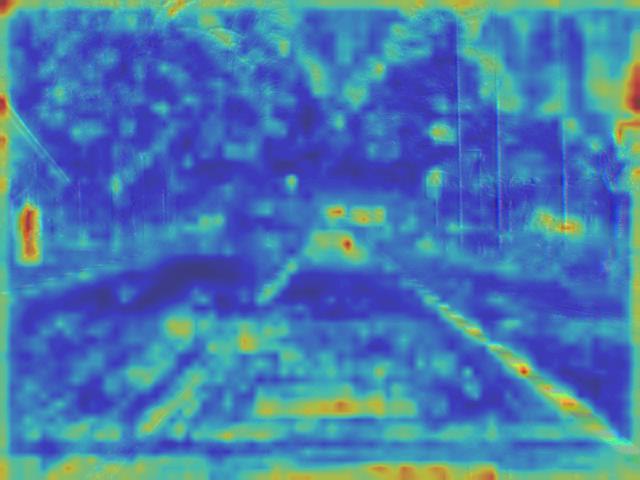}\\

\end{tabular}
\caption{Visualization of the spatial heat maps of the largest scale FPN feature in the YOLOX detector\cite{ge2021yolox} with different distillation methods on the DSEC dataset.}
\label{fig:comparison_dsec}
\end{figure*} 
\begin{figure*}[htbp]
\centering
\begin{tabular}{cccccc}
\rotatebox{90}{Grayscale Input} &\includegraphics[width=0.17\textwidth]{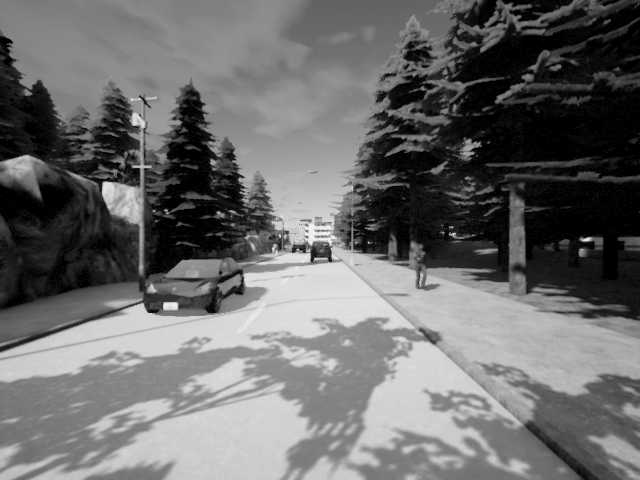}&
\includegraphics[width=0.17\textwidth]{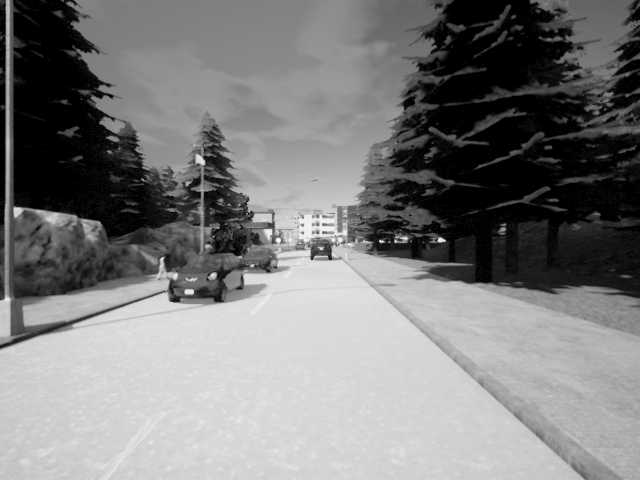}&
\includegraphics[width=0.17\textwidth]{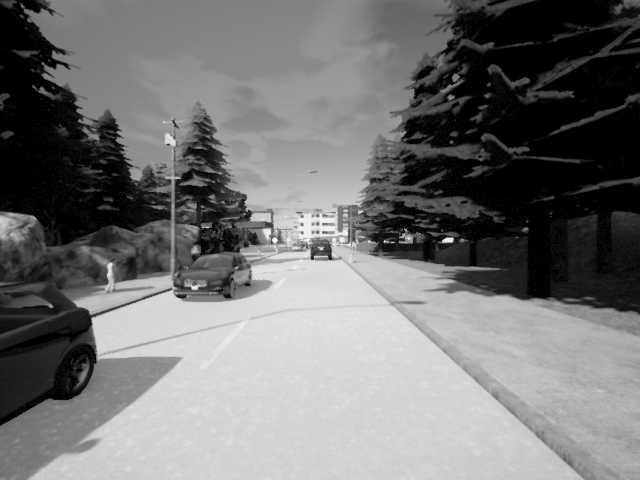}&
\includegraphics[width=0.17\textwidth]{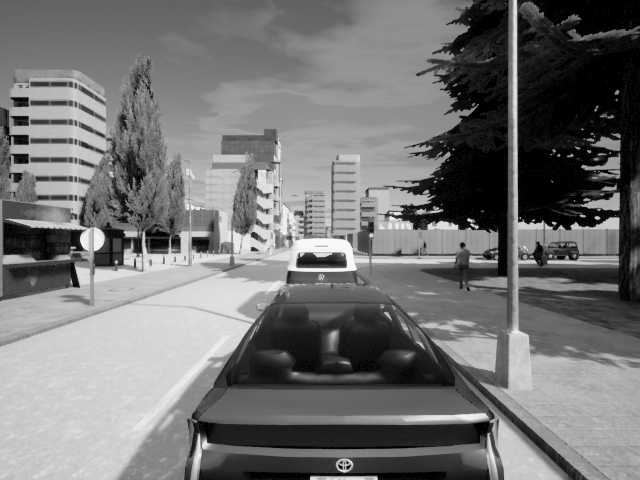}&
\includegraphics[width=0.17\textwidth]{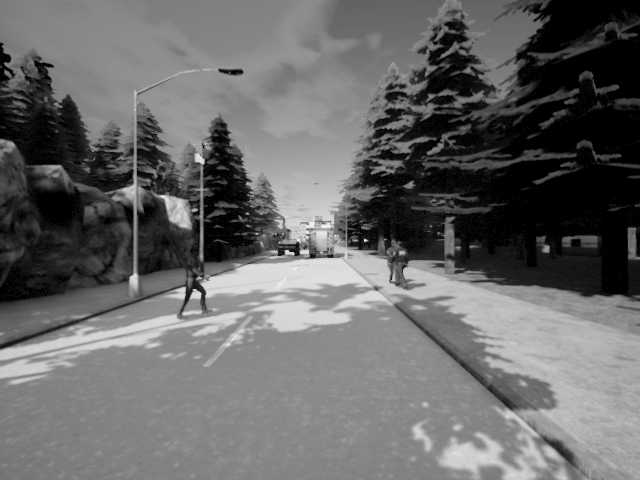}\\

\rotatebox{90}{~Events Input} &\includegraphics[width=0.17\textwidth]{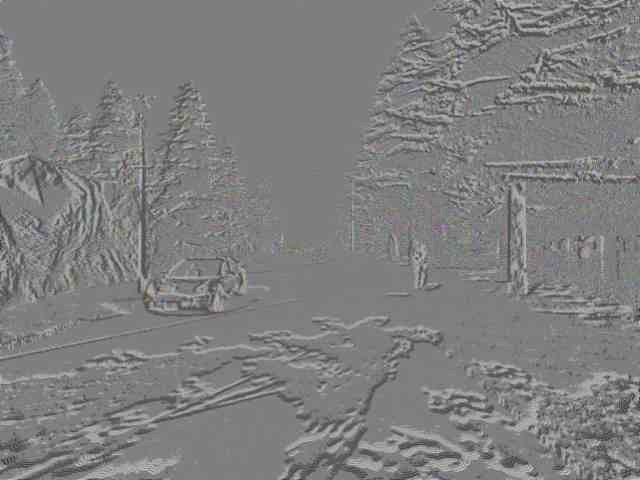}&
\includegraphics[width=0.17\textwidth]{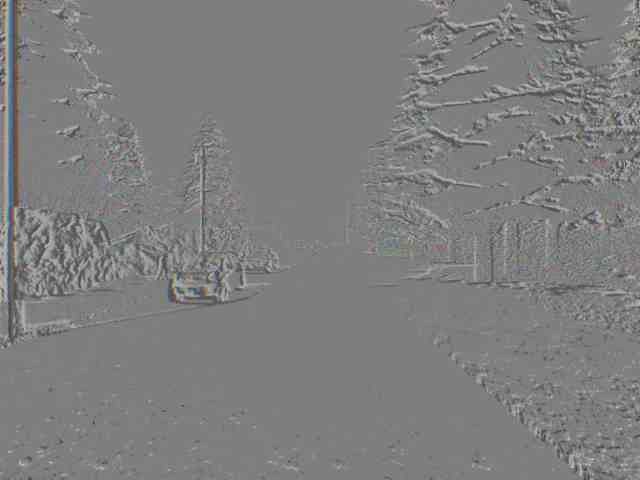}&
\includegraphics[width=0.17\textwidth]{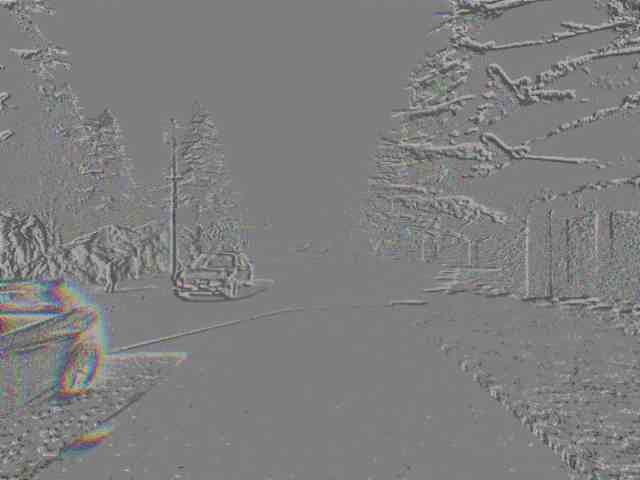}&
\includegraphics[width=0.17\textwidth]{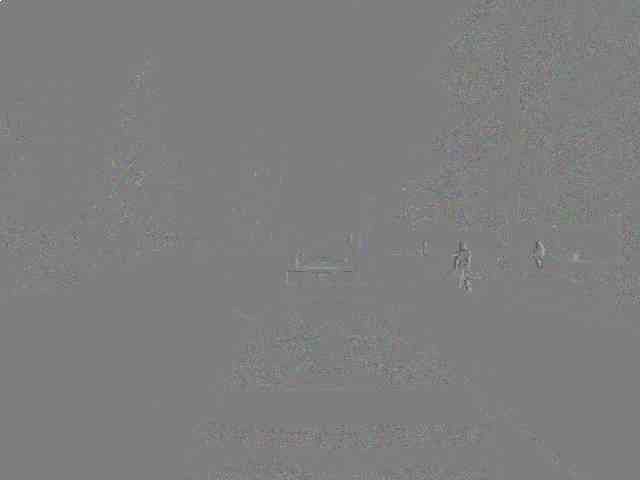}&
\includegraphics[width=0.17\textwidth]{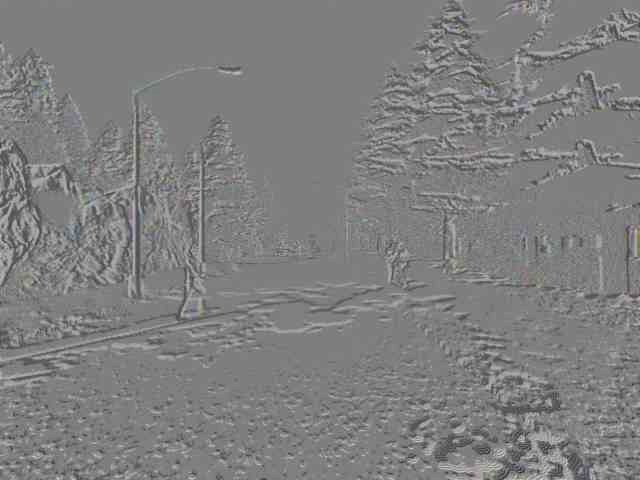}\\

\rotatebox{90}{~~~~~Grayscale} &\includegraphics[width=0.17\textwidth]{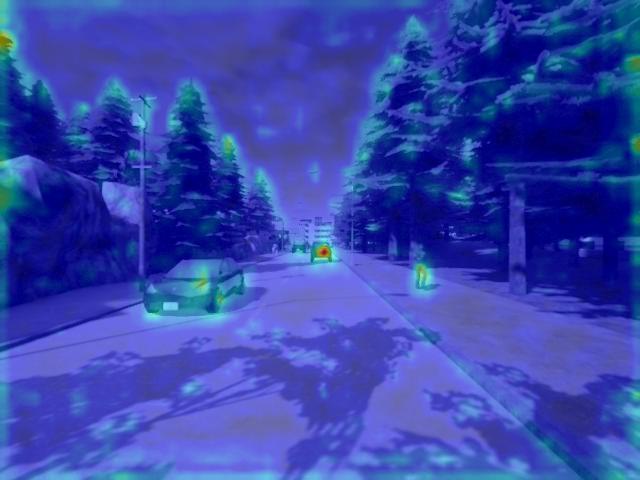}&
\includegraphics[width=0.17\textwidth]{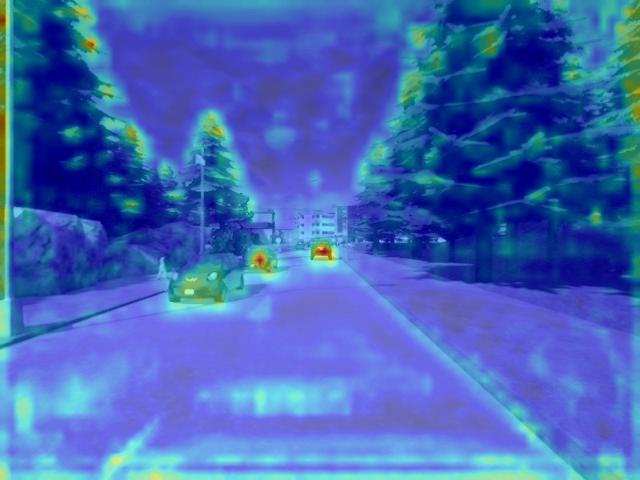}&
\includegraphics[width=0.17\textwidth]{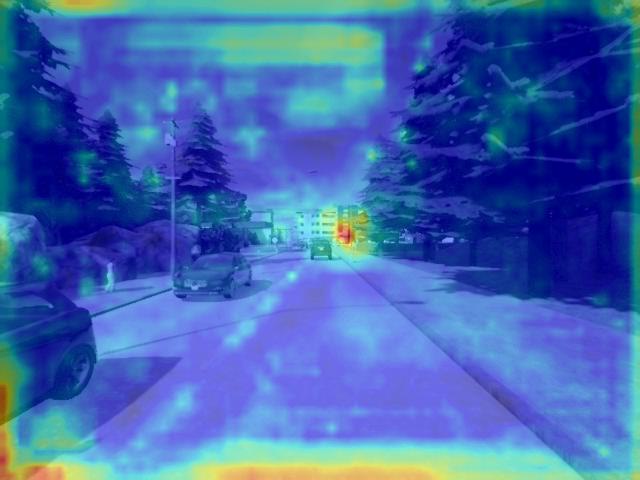}&
\includegraphics[width=0.17\textwidth]{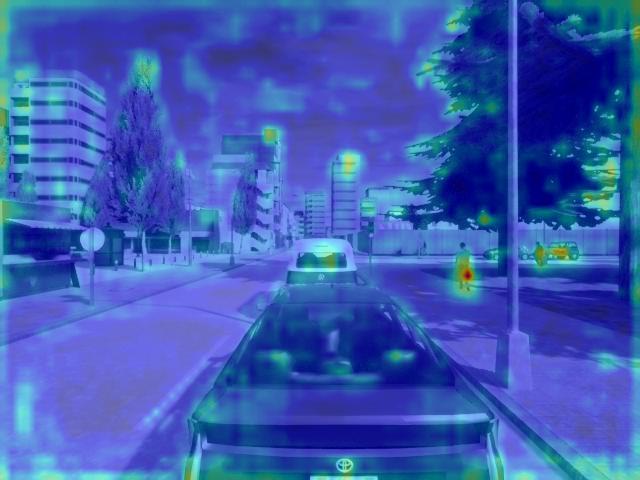}&
\includegraphics[width=0.17\textwidth]{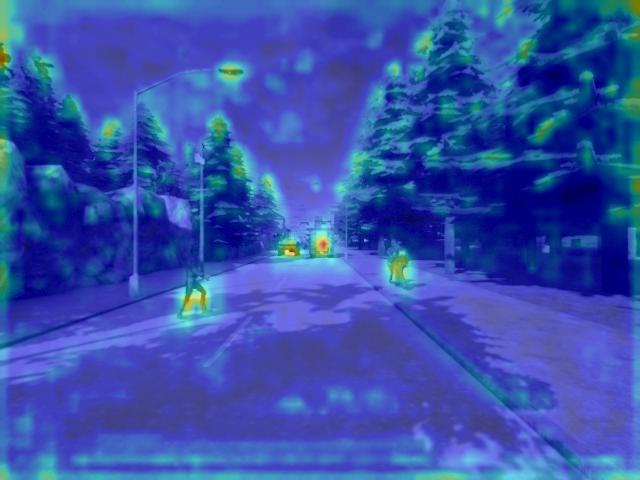}\\

\rotatebox{90}{~~~~~~~~Events} &\includegraphics[width=0.17\textwidth]{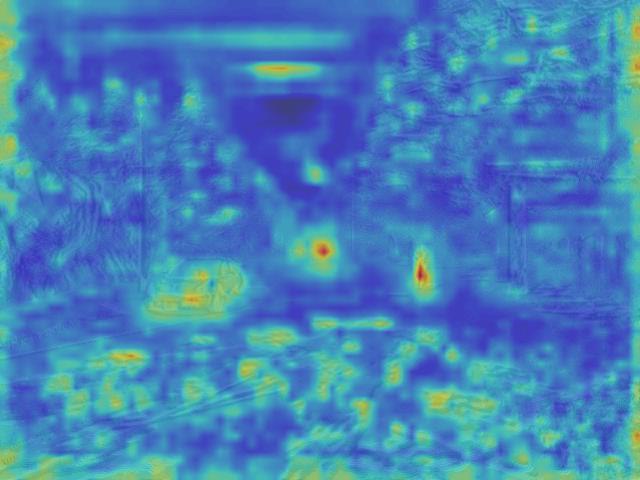}&
\includegraphics[width=0.17\textwidth]{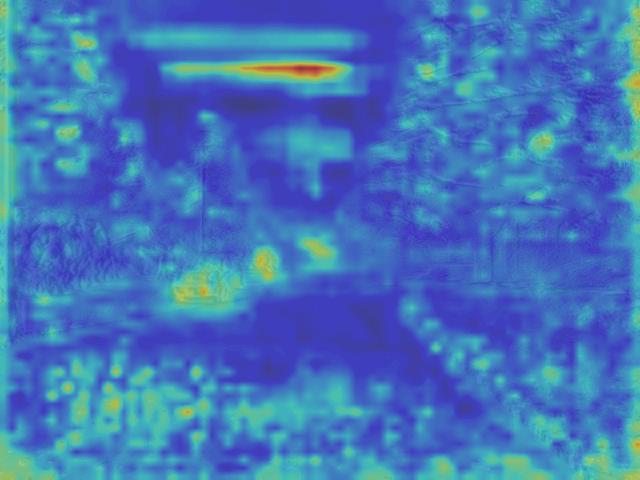}&
\includegraphics[width=0.17\textwidth]{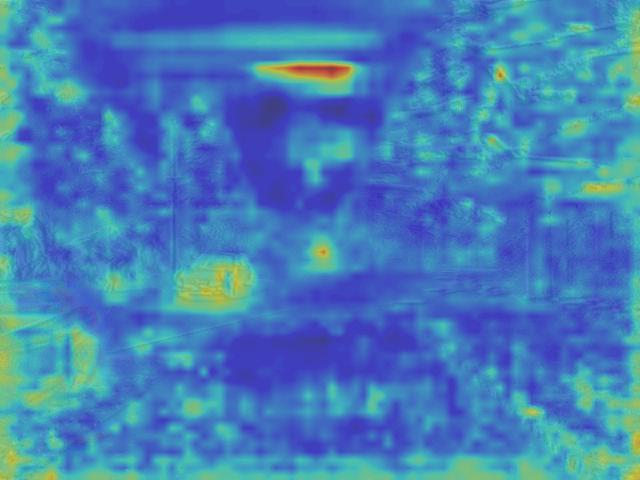}&
\includegraphics[width=0.17\textwidth]{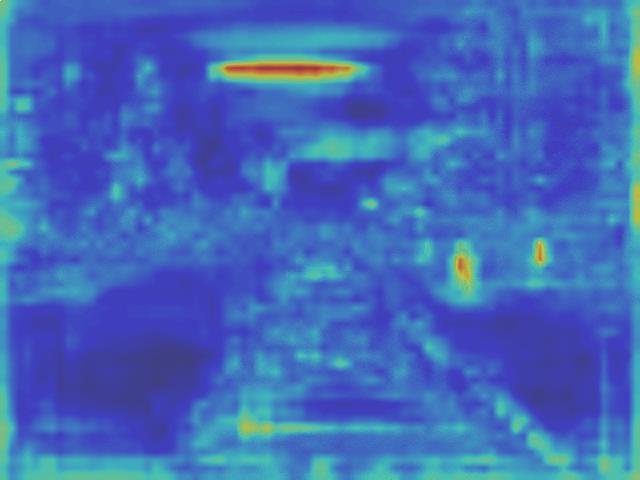}&
\includegraphics[width=0.17\textwidth]{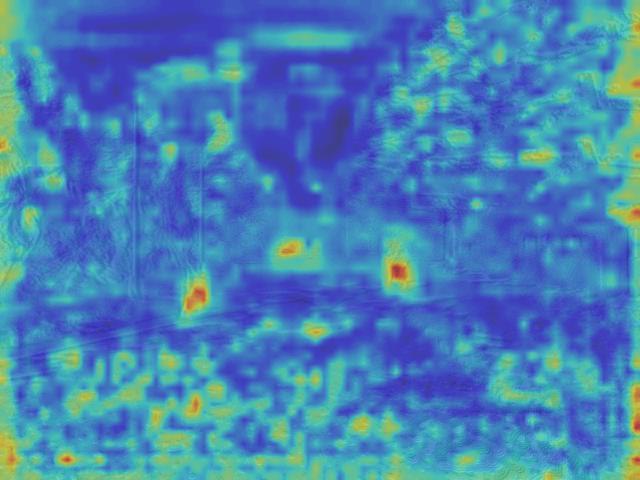}\\

\rotatebox{90}{~~~~~~~~+ICD} &\includegraphics[width=0.17\textwidth]{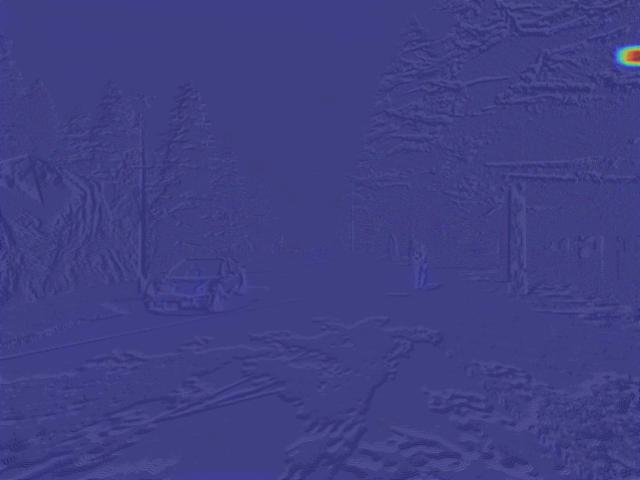}&
\includegraphics[width=0.17\textwidth]{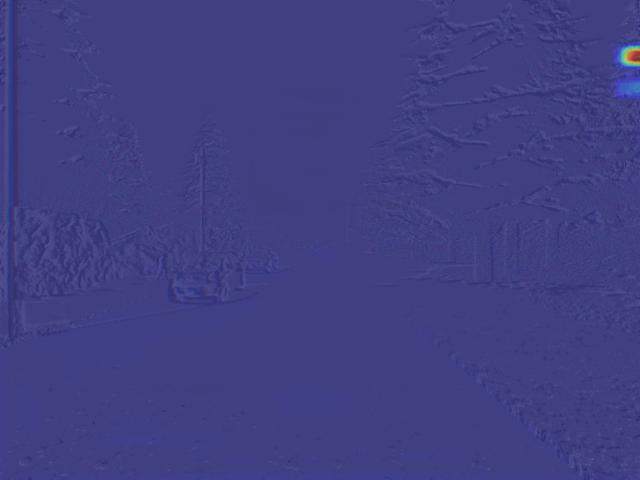}&
\includegraphics[width=0.17\textwidth]{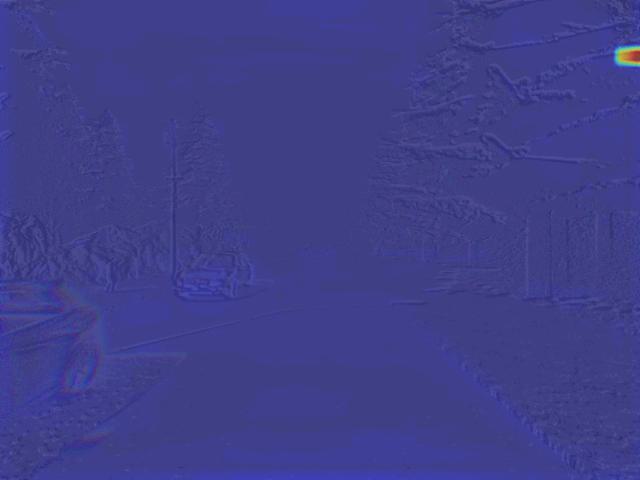}&
\includegraphics[width=0.17\textwidth]{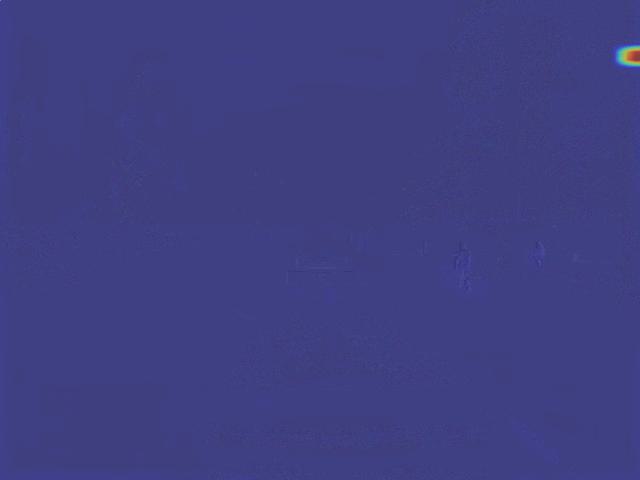}&
\includegraphics[width=0.17\textwidth]{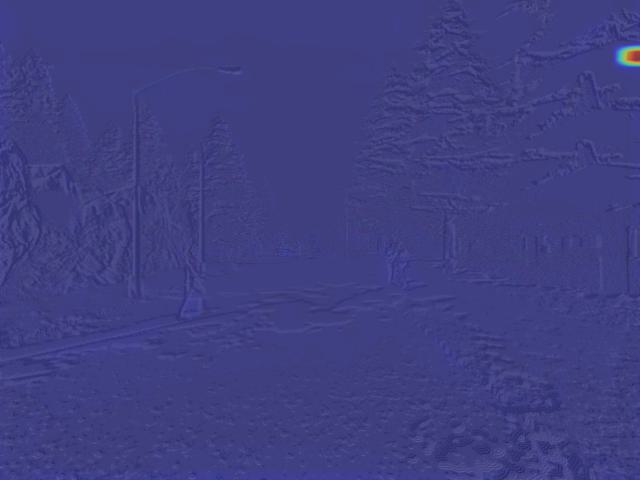}\\

\rotatebox{90}{~~~~~~~~+FGD} &\includegraphics[width=0.17\textwidth]{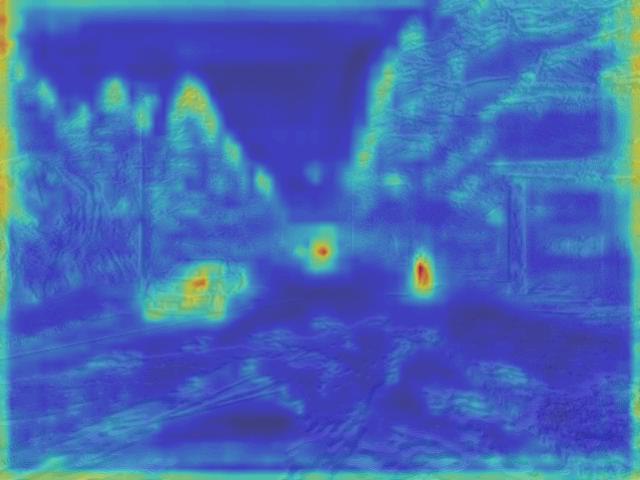}&
\includegraphics[width=0.17\textwidth]{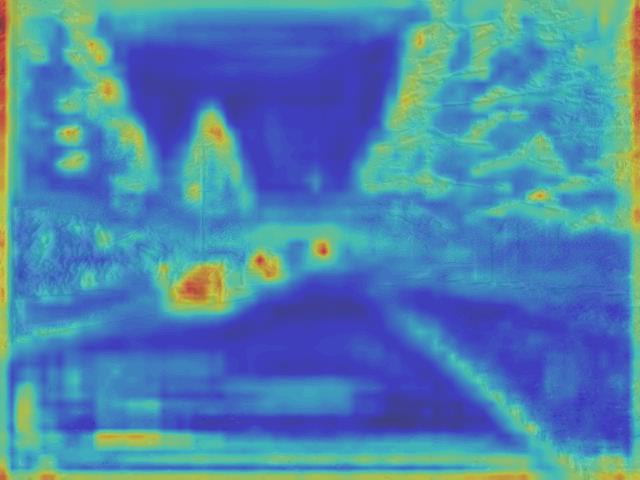}&
\includegraphics[width=0.17\textwidth]{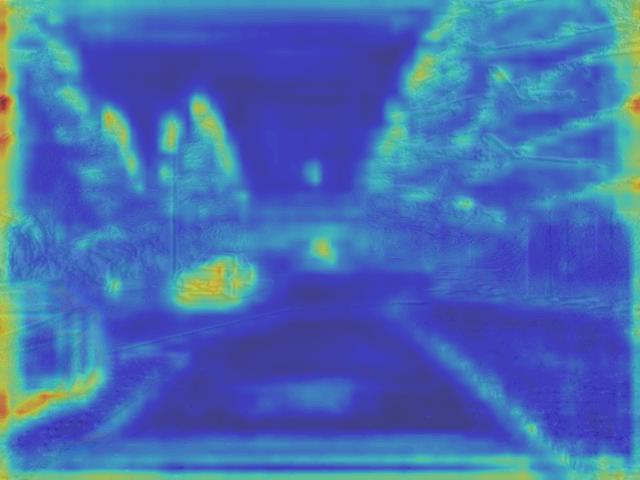}&
\includegraphics[width=0.17\textwidth]{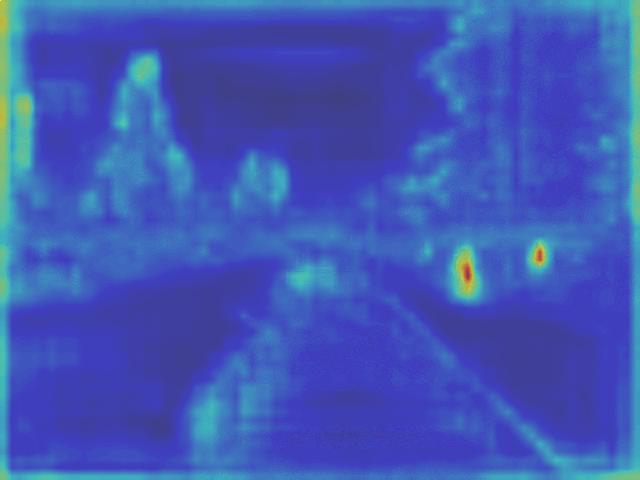}&
\includegraphics[width=0.17\textwidth]{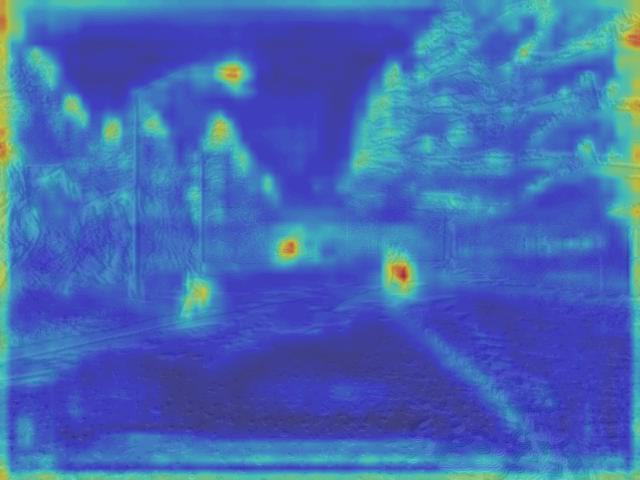}\\

\rotatebox{90}{~~~~~~~~+MGD} &\includegraphics[width=0.17\textwidth]{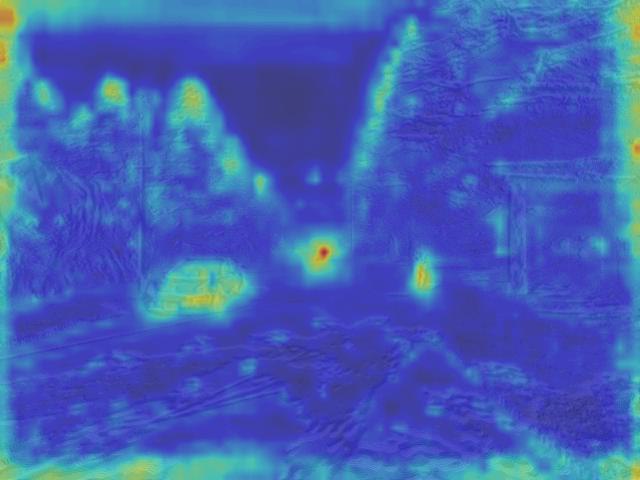}&
\includegraphics[width=0.17\textwidth]{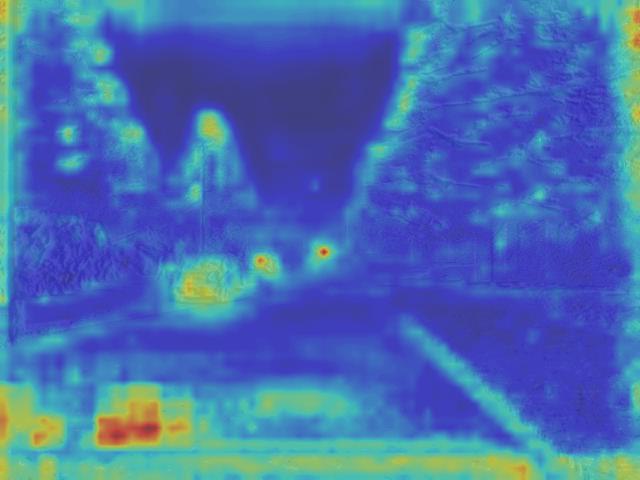}&
\includegraphics[width=0.17\textwidth]{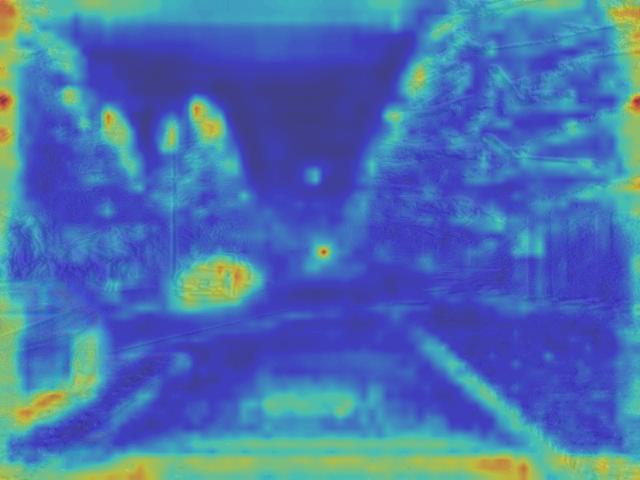}&
\includegraphics[width=0.17\textwidth]{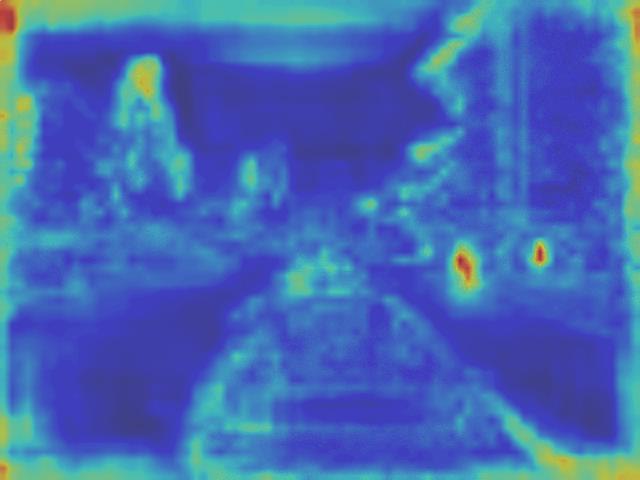}&
\includegraphics[width=0.17\textwidth]{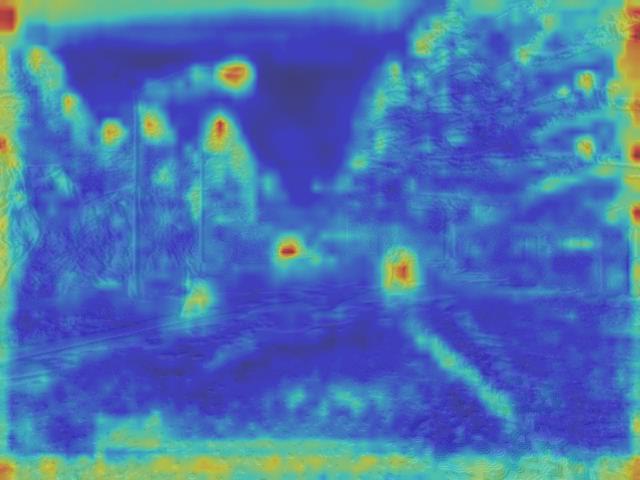}\\

\rotatebox{90}{~+MonoDistill} &\includegraphics[width=0.17\textwidth]{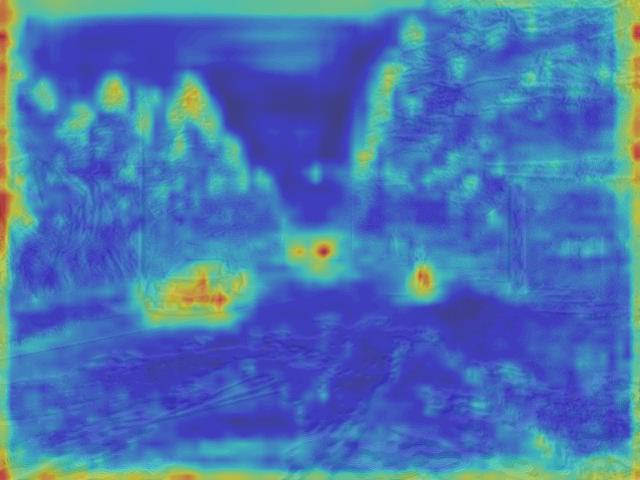}&
\includegraphics[width=0.17\textwidth]{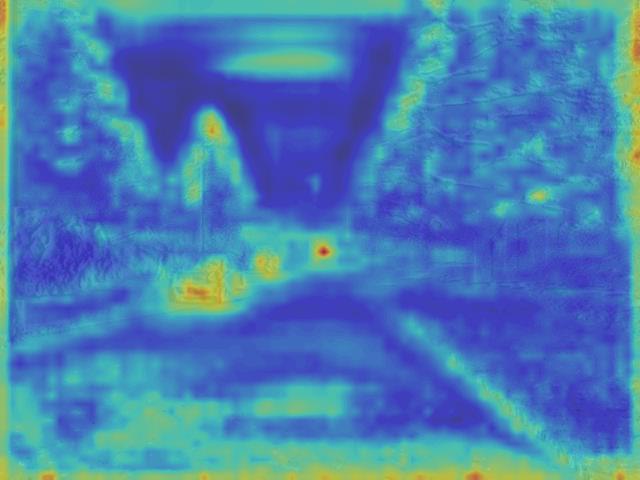}&
\includegraphics[width=0.17\textwidth]{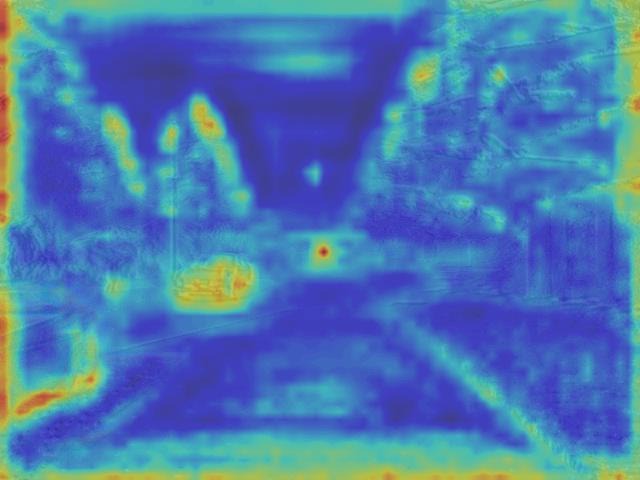}&
\includegraphics[width=0.17\textwidth]{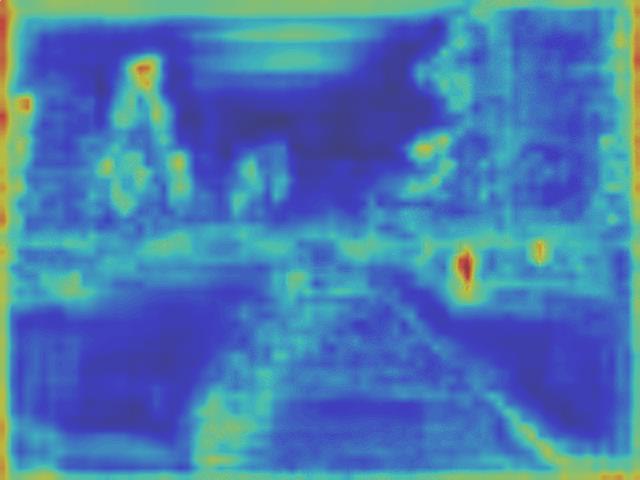}&
\includegraphics[width=0.17\textwidth]{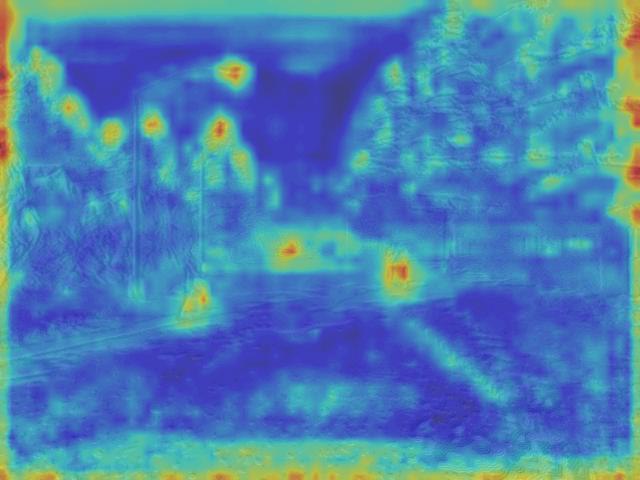}\\

\rotatebox{90}{~~~~~~~~~+Ours} &\includegraphics[width=0.17\textwidth]{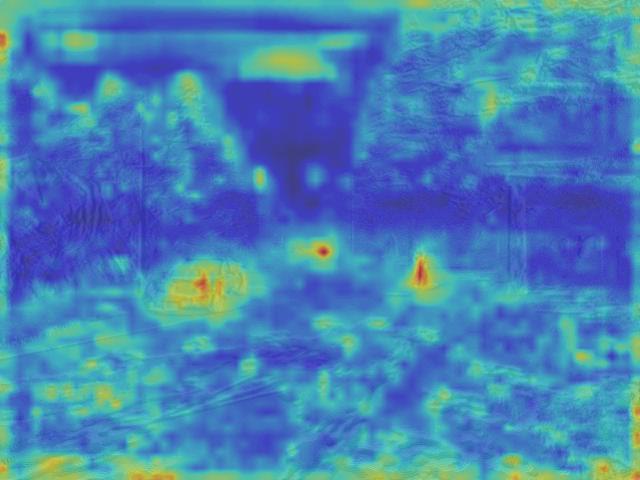}&
\includegraphics[width=0.17\textwidth]{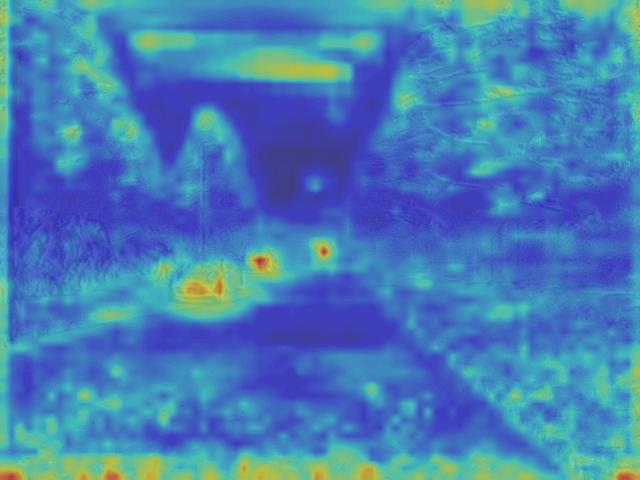}&
\includegraphics[width=0.17\textwidth]{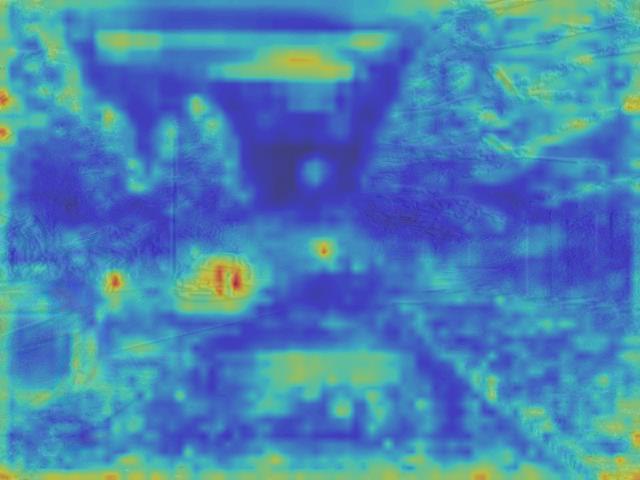}&
\includegraphics[width=0.17\textwidth]{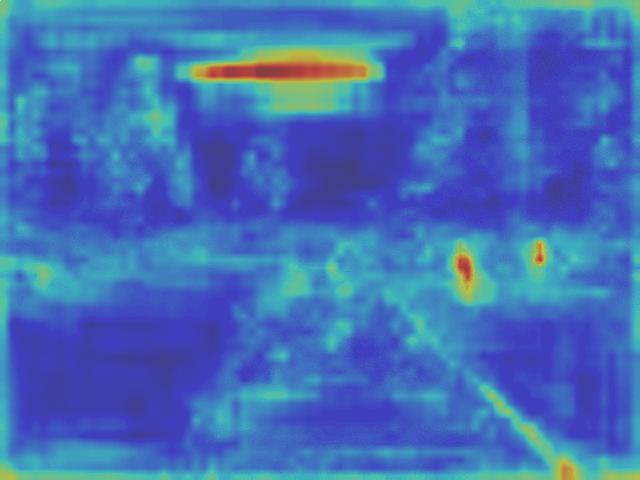}&
\includegraphics[width=0.17\textwidth]{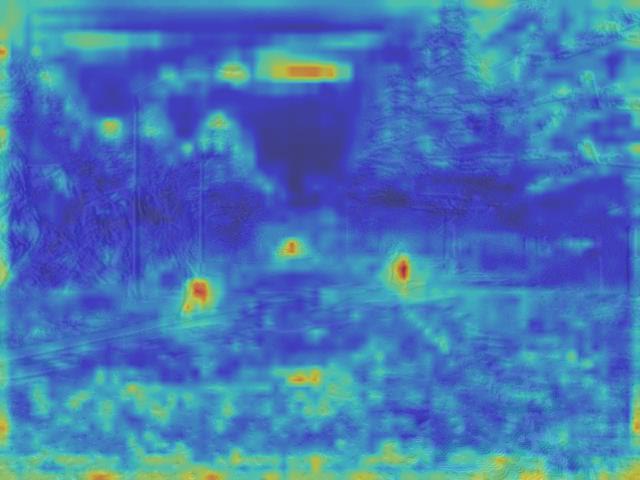}\\

\end{tabular}
\caption{Visualization of the spatial heat maps of the largest scale FPN feature in the YOLOX detector\cite{ge2021yolox} with different distillation methods on the CARLA dataset.}
\label{fig:comparison_carla}
\end{figure*}

\begin{figure*}[ht]
\centering
\scalebox{1.1}{
\begin{tabular}{cccc}
\includegraphics[width=0.2\textwidth]{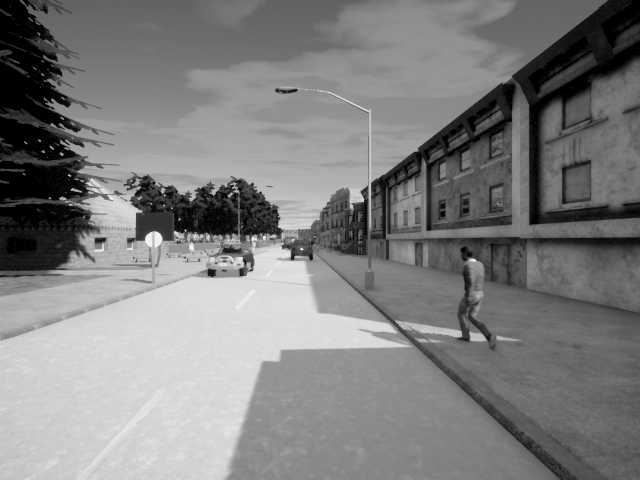}&
\includegraphics[width=0.2\textwidth]{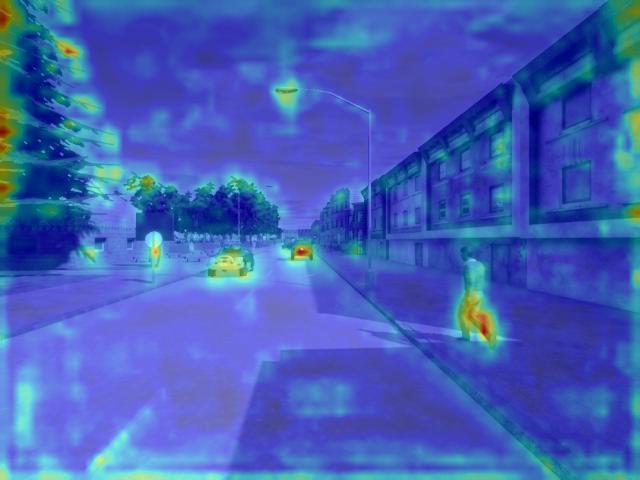}&
\includegraphics[width=0.2\textwidth]{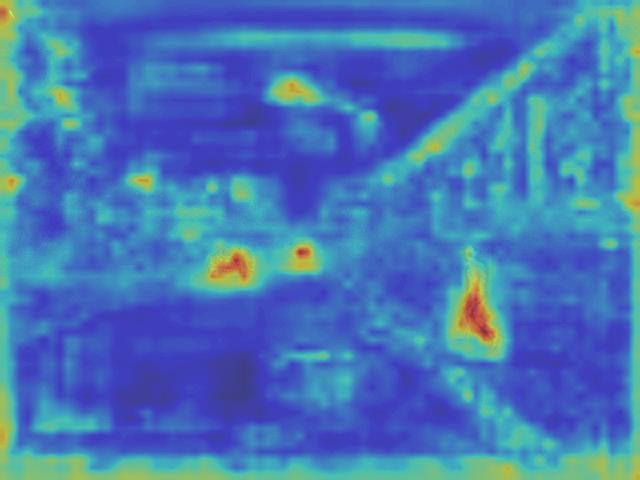} &  \includegraphics[width=0.2\textwidth]{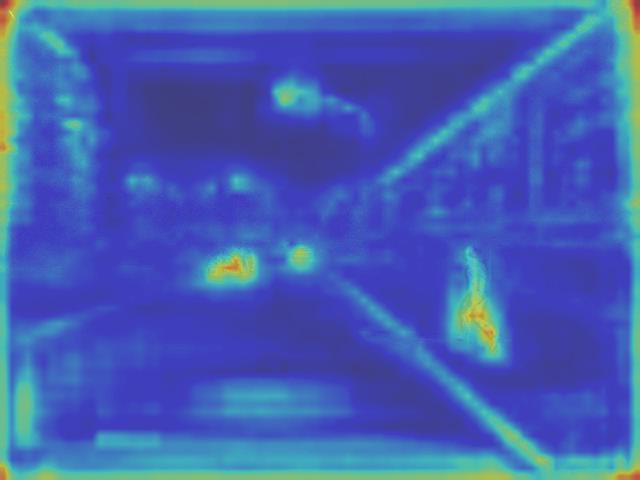} \\
Grayscale Input & Grayscale    &Event Frame    &+Full Region\\
\includegraphics[width=0.2\textwidth]{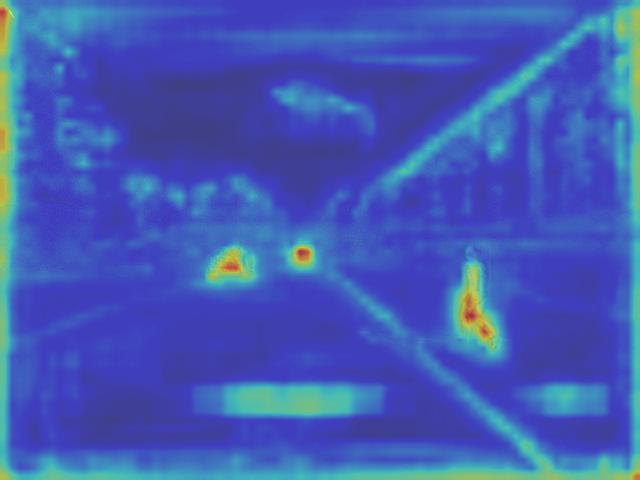}&
 \includegraphics[width=0.2\textwidth]{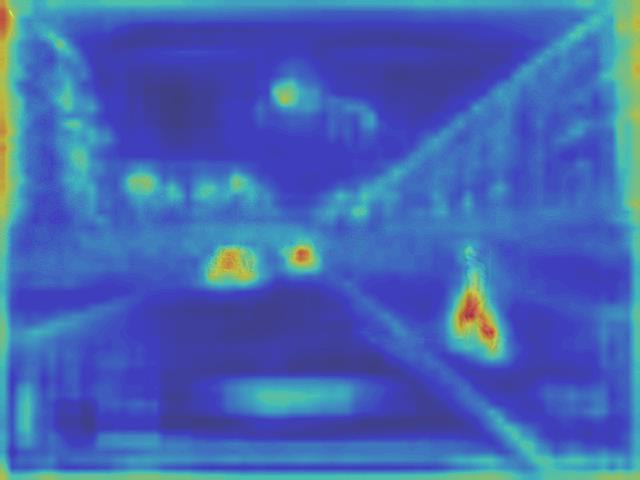}&
\includegraphics[width=0.2\textwidth]{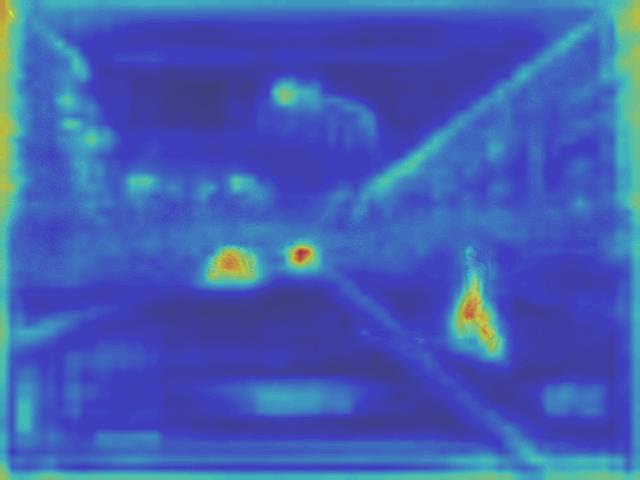}&
\includegraphics[width=0.2\textwidth]{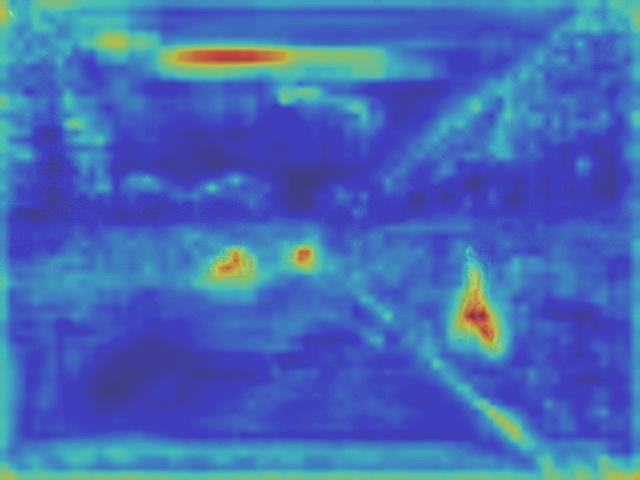}\\
+FG Region & +FGD    &+FGD FG    &+Ours \\
\end{tabular}
}
\caption{Visualization of the spatial heat maps of the FPN feature with different attention types.}
\label{fig:heatmap_attention}
\vspace{-0.4cm}
\end{figure*}

\subsection{More Details about Experiment Setting}
\noindent\textbf{More Dataset Details.} For DSEC, the validation set uses the following sequences, ``interlaken\_00\_e'', ``zurich\_city\_01\_e'', ``zurich\_city\_02\_e'', ``zurich\_city\_04\_e'', and ``zurich\_city\_09\_e'', which contain 6,903 labeled frames.

\noindent\textbf{Data Augmentation Details.} For all our experiments, we use random horizontal flips, and contrast and brightness changes as our data augmentation strategy. Our data augmentation pipeline simultaneously processes the grayscale image and event frame with the same probabilistic parameters. This works for geometric augmentations, such as scale jitter and flipping. However, even though we apply the same ``color" augmentations (brightness and contrast) these augmentations are not physically consistent between modalities, a change in the grayscale image would result in different events, but we keep them to increase data diversity.

\noindent\textbf{More Comparison Experiments Details.} For all state-of-the-art models in comparison experiments, we employ loss weights from their open-source code for corresponding one-stage and two-stage detectors. For our proposed method and ICD~\cite{kang2021instance}, auxiliary tasks are necessary to ensure that the feature extraction module does not degrade during training. Therefore, we train models with only original detection loss and auxiliary loss of the distillation module at the first epoch to stabilize feature extraction module. In the Faster R-CNN~\cite{ren2015faster} comparison experiment, we downscale the FPN feature dimension from 256 to 64 within the feature distillation module for our proposed method and ICD, ensuring that the distillation model size can fit into the limited GPU memory.

\noindent\textbf{More Implementation Details.} For our distillation module, $\mathcal{G}_c$ consists of two $3 \times 3$ convolutional layers and ReLU activation functions, and $\mathcal{G}_f$ consists of three $3 \times 3$ convolutional layers and ReLU activation functions. We train all detectors using SGD for 60 epochs, with a momentum of 0.937 and a weight decay of 0.0005. We use a cosine learning rate schedule with a 5 epoch warm up and use a learning rate of 0.000075 per sample in the batch. All experiments are run on a single Nvidia RTX 3090 GPU, with a batch size of 32 for YOLOX and 16 for Faster R-CNN. For all methods we first train the teacher model on the specific dataset using COCO pre-trained weights. Thus, the training of the teacher can be better understood as a fine-tuning step to grayscale images, utilizing the large scale COCO dataset as a starting point. For the event based student model, we also use COCO pre-trained weights, as we have seen that this improves detection performance.

\subsection{More Experiments}
\begin{table}[htbp]
\centering
\caption{\label{tab:ablation_modal} Ablation on teacher modality.}
\vspace{-0.2cm}
\scalebox{1.2}{
\centering
\begin{tabular}{c|lll}
\hline
Teacher  & mAP  & AP$_{50}$  & AP$_{75}$ \\ \hline
-  &53.8  &72.6  &58.4      \\ \hline
RGB  &62.7  &81.4  &72.7       \\
\rowcolor{Gray} + Ours  &51.1 (-2.7)  &68.8  &58.2\\\hline
Grayscale  &59.6  &81.1  &70.8\\      
\rowcolor{Gray} + Ours  &56.4 (+2.6)  &76.4  &63.2\\ \hline
\end{tabular}
}
\end{table}

\noindent\textbf{Teacher Modality.} 
The teacher modality can impact the student's performance since aligning feature works better the closer the two modalities are. Most frame based object detectors use RGB images as input, given the rich input, the network can focus on both texture and shape. However, event frames lack global illumination and texture details. Thus, event based detectors focus more on object shapes. We propose to use grayscale images as input for the teacher, the grayscale images do weaken the texture features of the response, bringing the features closer to the one of the event based detector. To validate this intuition, we perform an experiment where we compare RGB and grayscale teachers. In Table~\ref{tab:ablation_modal}, we can see that using RGB images results in a stronger teacher, with +3.1 mAP compared to grayscale. However, when using the RGB detector as a teacher, the performance is reduced by -2.7 mAP. This stands in contrast to our distillation from grayscale which improves the performance by 2.6 mAP. Thus, underlining our hypothesis that the added information in RGB images makes the distillation harder, as the two modalities are too different, especially their features.

\subsection{More Visualizations}
\noindent\textbf{Comparison Results on DSEC.}
We visualize the heat maps from the largest FPN feature map in some challenging scenarios, such as twilight and night. It is important to note that the largest FPN feature map exhibits a high response for small objects while having a weak response for large objects due to the limited receptive field. Figure \ref{fig:comparison_dsec} illustrates a qualitative comparison between our proposed method and state-of-the-art methods. For the first sample (first column), even in the presence of nighttime headlights, event-based detection is capable of reliably detecting distant vehicles without large disturbances. With our distillation approach, events' object feature responses are strengthened. Conversely, when grayscale images are used as input, the feature responses around vehicles become significantly weakened due to the influence of headlights. In the twilight scenario (second column), event-based detection with our distillation approach can detect pedestrians and distant cars. As for other distillation methods, ICD fails to localize objects, while FGD~\cite{yang2022focal} aligns background features with grayscale feature maps, leading to excessive smoothing of background regions and, consequently, a decrease in event-based detection performance. Although MGD~\cite{yang2022masked} and MonoDistill~\cite{chong2022monodistill} improve object feature responses, they introduce unwanted noise in the background area.

\noindent\textbf{Comparison Results on CARLA.} We also present visualization results on the CARLA dataset. Fig. \ref{fig:comparison_carla} showcases that the event data in the CARLA dataset share a similar form to the events in the DSEC dataset, despite the fact that the CARLA dataset is generated through the use of the CARLA Simulator~\cite{dosovitskiy2017carla}. With respect to the background regions, the event data contains numerous triggers, which are depicted as high responses in the FPN feature maps. Besides, we observe that all state-of-the-art methods show similar performance in the CARLA dataset as they act in the DSEC dataset.

\noindent\textbf{Attention Type.} 
To prove the experiment results in Sec.~\ref{sec:attention_type} and further highlight the differences between the approaches, we visualize the heatmap of the largest scale FPN features with distillation in Fig.~\ref{fig:heatmap_attention}. The feature map from event frames has a lot of active regions which correspond to the background, this is not the case for grayscale images. These active responses come from event triggers in noisy background regions, which have the same feature response principle as event triggers in foreground regions. Therefore, the intuition of our attention module is to maintain feature expressiveness on all event triggers and enhance feature expressiveness on foreground event triggers. As shown in Fig.~\ref{fig:heatmap_attention}, all baseline attention methods over-smooth the background feature responses and feature responses on foreground areas are weakened, which means these methods might damage feature expressiveness. In contrast, our approach strikes a balance between foreground and background feature responses.

\subsection{Limitations} 
Our current distillation approach relies on the availability of paired event and image datasets during training, which is a limiting factor. The requirement for paired event-image data is only present for training and not for deployment. Relying on auto-labeling for real event data is another limitation. We hope to alleviate these limitations by extending our method to work with unlabeled data. 

\subsection{Future Work} 
Our approach enhances event-based object detection performance by transferring knowledge from images to event frames. In Section~\ref{sec:generalization}, we demonstrate that our method can also be applied in the conventional distillation setting, where knowledge is transferred from a large model to a small model, highlighting the versatility of our approach across different experimental scenarios. As sensor fusion technology continues to evolve, another promising future work is extend our work into distilling knowledge from fused event frames and images into images, or from fused frames and images into event frames.

\end{document}